\documentclass[fleqn,11pt]{article}

\usepackage{multicol}
\usepackage{multirow}
\usepackage{makecell}
\usepackage{array}
\usepackage{amsmath}
\usepackage{ctable}
\usepackage{enumitem}
\usepackage{pifont}
\newcolumntype{P}[1]{>{\centering\arraybackslash}p{#1}}

\usepackage[para,online,flushleft]{threeparttable}
\usepackage{soul}
\usepackage{sidecap}
\usepackage{epsfig}
\usepackage{color}
\usepackage{graphicx}
\usepackage{multirow}
\usepackage{hanging}
\usepackage{wrapfig}
\usepackage{subfig}
\usepackage{epstopdf}
\usepackage{float}
\usepackage{booktabs}
\usepackage[small,compact]{titlesec}
\usepackage{palatino}
\usepackage{xurl}
\usepackage[rflt]{floatflt}
\usepackage{subfig}
\usepackage{comment}
\usepackage{amssymb}
\usepackage{multirow}
\usepackage{cite}
\usepackage{tabularx}
\usepackage{xcolor}
\usepackage{listings}
\usepackage{mscp}
\graphicspath{{./figures/} }
\usepackage{multicol}
\usepackage{threeparttable}

\definecolor{dkgreen}{rgb}{0,0.6,0}
\definecolor{gray}{rgb}{0.5,0.5,0.5}
\definecolor{mauve}{rgb}{0.58,0,0.82}

\newcounter{myctr}

\newcommand{\ignore}[1]{}
\newcommand{\rv}[1]{#1}

\usepackage{xspace}
\newcommand{\projectname}{XGen\xspace}

\title{
\vspace*{-\baselineskip}
\vspace*{-0.8\baselineskip}
\Large \bf
CoCoPIE XGen: A Full-Stack AI-Oriented Optimizing Framework
}
\author{Xiaofeng Li, Bin Ren, Xipeng Shen, Yanzhi Wang\\
Contact: info@cocopie.ai}

\date{}

\begin{document}


\maketitle
\vspace*{-0.5\baselineskip}


\pagestyle{plain}


\section{Introduction}

\begin{figure}
\centering
\includegraphics[width=.8\textwidth]{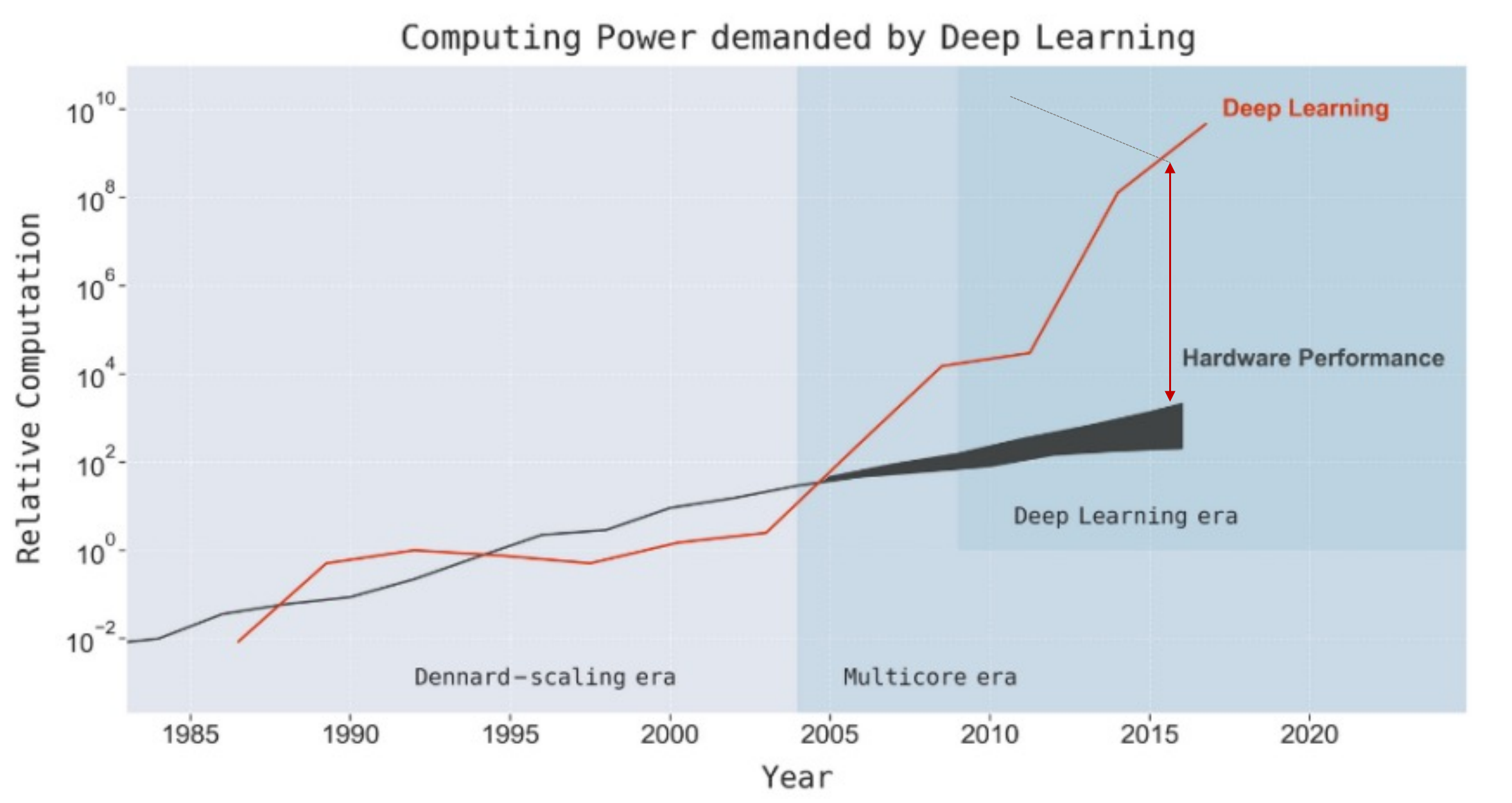}
\caption{Growing gap between the computing demands of DNNs and the computing power offered by modern hardware including AI accelerators.~\cite{thompson2020computational}}\label{fig:gap}
\end{figure}

Recent years have witnessed a rapid rise of Deep Learning. This Deep Neural Networks (DNN) based technology has given drastically improved results over traditional machine learning on a wide range of Artificial Intelligence (AI) tasks. It has quickly become the new norm of AI. Its recent development has shown two important trends. First, the size of DNNs and their demands for computing power have grown in a much higher rate than that of the underlying computing hardware, as shown in Figure~\ref{fig:gap}. Second, there is a growing demand for shifting the delivery of AI capability from data centers on the cloud to edge or end devices, exemplified by the fast emerging real-time AI-based apps running on smartphones, AR/VR devices, autonomous vehicles, and various IoT devices. The shift has however been seriously hampered by the even larger gap between DNN computing demands and the computing power on edge or end devices. We call the gap {\em DNN-hardware speed gap}.

To fill the DNN-hardware speed gap, software solutions are essential. Even though many software tools (e.g., PyTorch \cite{pytorch}, TensorFlow \cite{abadi2016tensorflow}, MNN \cite{ali-mnn}, TVM \cite{chen2018tvm}, TensorRT \cite{tensorRT}) have been created in the recent years to tackle the issue, their effectiveness has been seriously limited by a principled shortcoming in their designs, namely {\em siloed optimization}. To bridge the huge gap illustrated in Figure~\ref{fig:gap}, optimizations have to happen at every layer in the software stack, from the DNN model to its computation flow, code generation, deployment and execution. More importantly, as the results of these optimizations affect each other, their full potential cannot be unlocked unless they are designed together in a hand-in-hand manner. None of the existing frameworks are doing that. Some of them may cover optimizations on more than one layer of the software stack, but they are designed separately, or, at the very best, in an only loosely related manner. 

XGen from CoCoPIE is created to fill the void. XGen is an optimizing framework for DNN. It takes cross-cutting co-design as its first-order consideration. Its {\em full-stack AI-oriented optimizations} consist of a number of innovative optimizations at every layer of the DNN software stack, all designed in a cooperative manner. The unique technology makes XGen able to optimize various DNNs, including those with an extreme depth (e.g., BERT, GPT, other transformers), and generate code that runs several times faster than those from existing DNN frameworks, while delivering the same level of accuracy. This article provides an overview of the core technology inside XGen, its product form, the comparisons with other DNN optimizing frameworks, and the demonstrations of its use in several real-world applications. (XGen is potentially useful for accelerating both the training and inferences of DNNs, although currently its main focus is DNN inference.)

\section{Technology inside XGen} 

\begin{figure}
    \centering
    \includegraphics[width=0.99\textwidth]{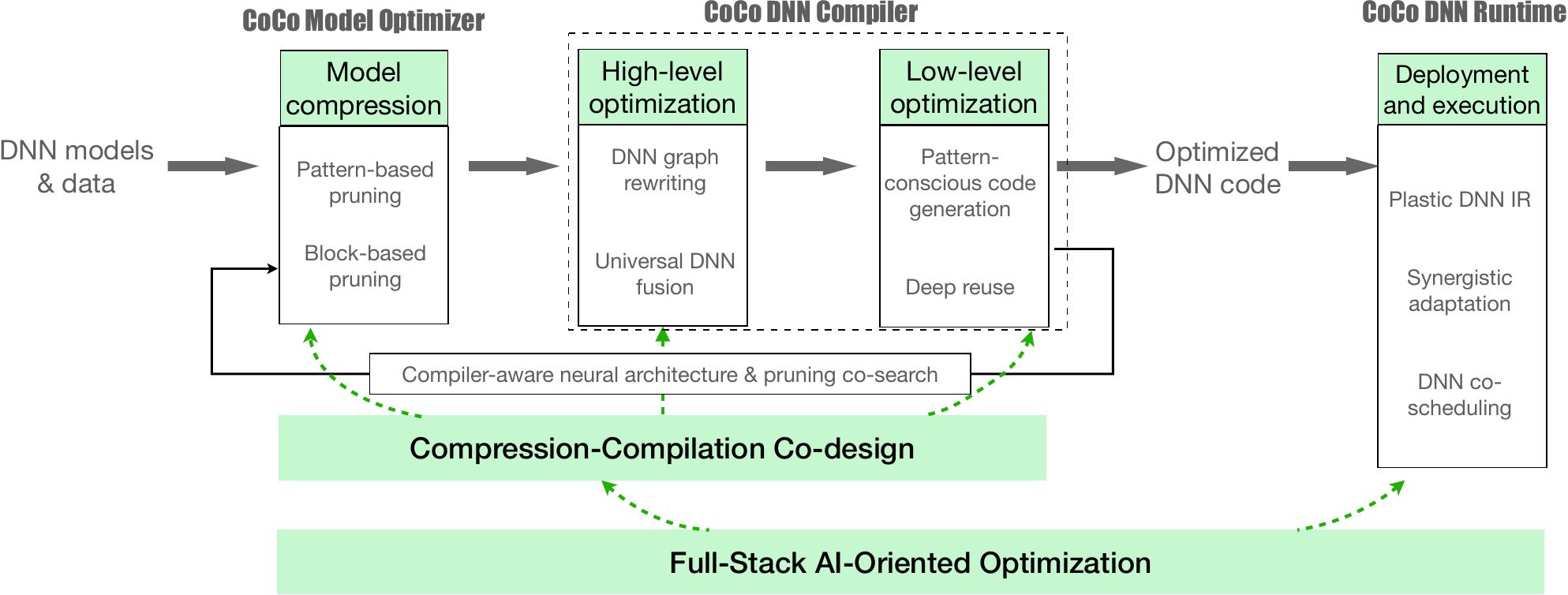}
    \caption{High-level view of the core technology of XGen.}
    \label{fig:xgen}
\end{figure}

Figure~\ref{fig:xgen} provides an overview of the key components of XGen and their roles in optimizing DNN. As a full-stack optimizing framework for DNN, it optimizes a given DNN model at all layers of the stack. 

\begin{itemize}

   \item For a given DNN model written in some common DNN APIs (e.g., PyTorch, TensorFlow), XGen first compresses the model through {\bf CoCo model optimizer}, which reduces the size and complexity of the DNN model in a manner friendly to the later optimizations via {\em pattern-based pruning} and {\em block-based pruning}, along with the compatible model compression techniques, such as quantization and knowledge distillation. 

   \item The optimized DNN model then goes through {\bf CoCo DNN compiler}, which consists of two-level code optimizations. The {\em high-level optimization} streamlines the DNN computations at the level of DNN graphs through {\em DNN graph rewriting} and {\em universal DNN fusion}. The {\em low-level optimization} ensures efficient (parallel) code being generated through {\em pattern-conscious code generation}~\cite{niu2020patdnn} and {\em deep reuse}. The output of the compiler is executable code that implements the DNN model for inference. 

   \item The deployment of the generated DNN code and its execution on devices can be (optionally) further optimized by {\bf CoCo DNN runtime}, a lightweight runtime system that coordinates DNN model deployment and optimizes resource utilization in the presence of multiple DNN tasks. The runtime does it effectively through {\em plastic DNN IR}, {\em synergistic adaptation}, and {\em DNN co-scheduling}. 
\end{itemize}

{\em Compression-compilation co-design} ties the first three components in the flow together. It reduces the DNN model size while preserving regular patterns and/or blocks in DNN kernels. It enables co-optimizations of the model architecture and pruning with code generation through an innovative {\em compiler-aware neural architecture \& pruning co-search (CAPS)}. As the patterns and blocks are designed hand-in-hand with the compiler, the optimized DNN model exhibits a form best fit for the compiler to generate efficient code for the target device. Both the high-level and low-level optimizations in the compiler take advantage of the regular patterns, removing as much redundant computation as possible while emitting code that fully taps into the underlying parallel computing units and memory hierarchy. The last component, CoCo runtime, is designed to be conscious of the effects of DNN model and code optimizations. It creates the optimal schedules for a set of DNN models and other modules with complicated dependence, such that these tasks can progress smoothly in a resource-constrained environment while meeting the expected quality of service (QoS). It supports devices equipped with heterogeneous computing units (e.g., CPU, GPU, DSP, DLA). It achieves that by distinctively co-optimizing the DNN models and schedules, and combining offline AI model analysis and just-in-time priority adjustment.

XGen builds on a set of proprietary technology, invented in the 20+ combined years of research by the three world-class research groups affiliated with CoCoPIE (Northeastern University, The College of William \& Mary, and North Carolina State University), and protected by over ten patents. The principle of co-design threads the designs of all the components in XGen. We next explain each of the components. 



\subsection{Model Optimization}
\label{sec:cocogen}

\begin{figure}
    \centering
    \includegraphics[width=0.5 \textwidth]{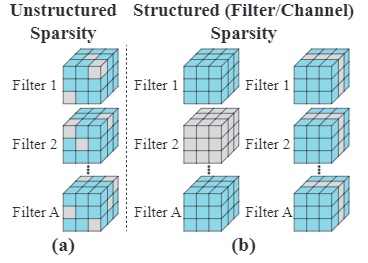}
    \caption{(a) Non-structured weight pruning and {(b) two types of structured weight pruning.}}
    \label{fig:structuredpruning}
\end{figure}

The first component of XGen optimizes DNN models. Its main goal is to reduce the size and complexities of the input DNN model. It does it through unique DNN weight pruning techniques named {\em pattern-based pruning} and {\em block-based pruning}, and uses a {\em composability-driven method} to minimize the pruning time even in the presence of an enormous pruning space. XGen is also compatible with orthogonal DNN compression techniques such as quantization and knowledge distillation.

{\em DNN weight pruning} is one of the most effective ways to reduce the size of a DNN and its computations. Prior work on DNN pruning falls into three categories. 




{\em 1) Non-Structured Pruning.}
Figure~\ref{fig:structuredpruning}(a) illustrates this method, where arbitrary weights can be pruned. It can give a high pruning rate (i.e., reduction in the number of weights) without degrading the accuracy. However, for compiler and code optimization, non-structured pruning incurs several challenges due to the irregularity in computation and memory access. Similarly, for hardware acceleration, since the pruned models are stored in some sparse matrix format with indices, they often lead to performance degradation in GPU and CPU implementations \cite{wen2016learning,he2017channel,mao2017exploring}. 

{\em 2) Structured Pruning.}
This method can produce smaller regular weight matrices.
{Figure \ref{fig:structuredpruning} (b)} illustrates the typical structured pruning schemes: \emph{filter pruning} and \emph{channel pruning} \cite{wen2016learning}. 
Filter and channel pruning can be considered as equivalent in that pruning a filter in the $k$-th layer is equivalent to pruning the corresponding channel in the $(k+1)$-th layer. Filter/channel pruning is compatible with Winograd algorithm \cite{winograd1980arithmetic,lavin2016fast} 
that has been used to accelerate 
computation of the original DNNs. 
Due to the regular structure, 
the GPU/CPU implementations typically lead to more significant acceleration \cite{wen2016learning,he2017channel}.
However, the structured pruning suffers from notable accuracy loss~\cite{wen2016learning,he2017channel}.

{\em Other Types of Weight Pruning.} There are a few other types of DNN pruning schemes (e.g., vector-based~\cite{mao2017exploring} and those supported by PyTorch and TensorRT) that produce sparsity between the above definitions of non-structured and structured pruning. These pruning schemes aim to find a balance between the sparsity degree, the computation load, and the achieved accuracy. These techniques essentially can be categorized into structured and non-structured pruning. They either are not well compatible with the underlying vector machine of general-purpose computing devices, or cannot well preserve the original model accuracy. 


To address the limitations of the prior methods, XGen develops two hardware-aware compiler-friendly DNN pruning methods, namely {\em pattern-based pruning} and {\em block-based pruning}. The former gives the best known results on a class of DNNs. The latter is a generalization of the former; it keeps most of the benefits of the former, and at the same time, extends the applicability to all kinds of DNNs. 

\subsubsection{Pattern-Based Pruning}

{\em Pattern-based pruning} achieves the high accuracy of non-structured pruning and the hardware friendliness of structured ones. It does that by creating \emph{fine-grained pruning patterns inside the coarse-grained structures}.

\begin{figure}
    \centering
    \includegraphics[width=0.65 \textwidth]{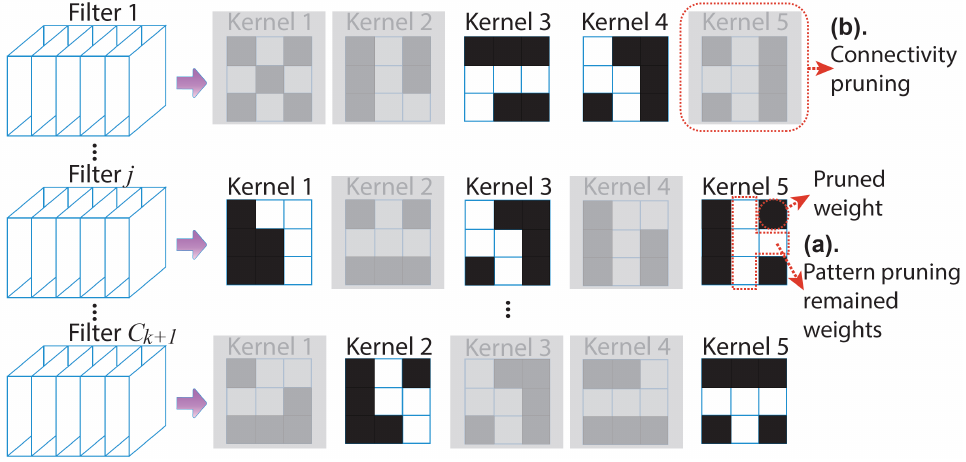}
    \caption{Illustration of (a) kernel pattern pruning on CONV kernels, and (b) connectivity pruning by removing kernels.}
    \label{fig:pattern_connectivity}
\end{figure}

Figure \ref{fig:pattern_connectivity} illustrates the basic idea of {\em pattern-based pruning}. For each kernel (in a CONV filter), a fixed number of weights are pruned,
and the remaining weights (white cells) form specific ``patterns''. We define the example in {Figure \ref{fig:pattern_connectivity}} as 4-entry pattern pruning, since every kernel reserves 4 non-zero weights out of the original $3\times 3$ kernel (the most commonly used kernel). It can generalize to other kernel sizes and fully connected layers. Each kernel has the \emph{flexibility} in choosing among a number of pre-defined patterns.

At theory and algorithm levels, such patterns exhibit similarities to the connection structure in human visual systems~\cite{ma2019pconv,li2017pruning,lebedev2016fast}. At compiler level, the known patterns allow a compiler to \emph{re-order and generate codes} at filter and kernel level such that kernels with the same pattern can be grouped together for consecutive executions, thereby maximizing instruction-level and thread-level parallelism. At hardware level, 4-entry patterns perfectly fit the SIMD architecture in embedded processors, for both CPUs and GPUs.

The selection of appropriate patterns for a kernel can be achieved via search through an extended ADMM-based framework~\cite{ma2019pconv}.

The method can be used together with {\em connectivity pruning}, which \emph{cuts the connections} between certain input and output channels, to achieve even higher weight pruning/acceleration rates. 

Although {\em pattern-based pruning} gives the best results on a class of DNNs where the DNN kernel size is among a set ($3\times3$, $5\times5$, $7\times7$), the number of pattern candidates grows rapidly as the kernel size increases, which exacerbates the computation irregularity and degrades the execution performance. The limitation prompts the development of {\em block-based pruning}.   

\subsubsection{Block-Based Pruning and Generalization to 3D Convolutions}

\begin{figure}
    \centering
    \includegraphics[width= \textwidth]{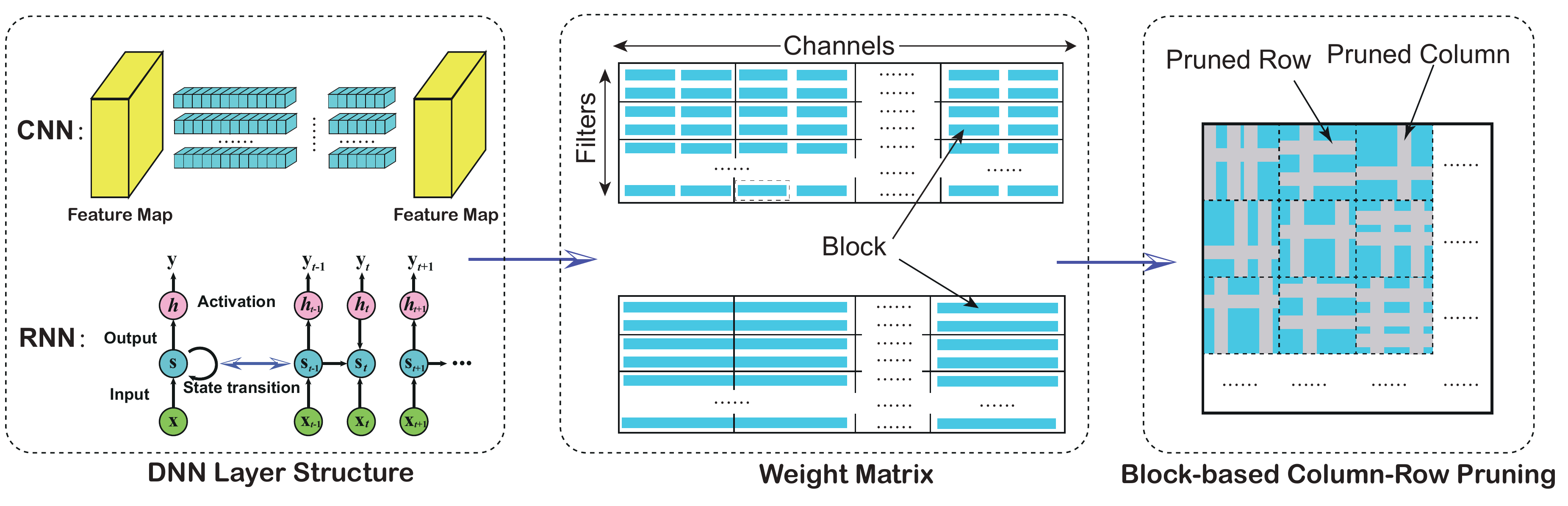}
    \caption{Illustration of block-based pruning that applies to all of CNN, RNN, transformers and to different layer types.}
    \label{fig:block_pruning}
\end{figure}

As an effective complement to the above pattern-based pruning, we develop \emph{block-based pruning} \cite{niu2021grim,cai2021yolobile} that is a general pruning scheme that applies to all of CNNs, RNNs, transformers and all types of network layers. 
Specifically, for any weight matrices in CNNs and RNNs, we first partition it into a number of \emph{weight blocks} and then apply independent column pruning and row pruning to each block, as shown in Figure \ref{fig:block_pruning}. 
Please note that the operations in CONV layers can be transformed into the general matrix multiplication (GEMM) routine \cite{chetlur2014cudnn} and therefore we can obtain the corresponding matrix format for filters in a CONV layer. We have extended the ADMM-based pruning algorithm to automatically determine the block-based sparsity for each block and each network layer, as well as a algorithm-compiler co-design to determine the appropriate, layerwise block size.
Our block-based pruning enjoys simultaneously the high accuracy as the non-structured sparsity and the regularity as the course-grained structured sparsity, achieving all their advantages while overcoming their shortcomings.

\begin{figure}
    \centering
    \includegraphics[width=0.45 \textwidth]{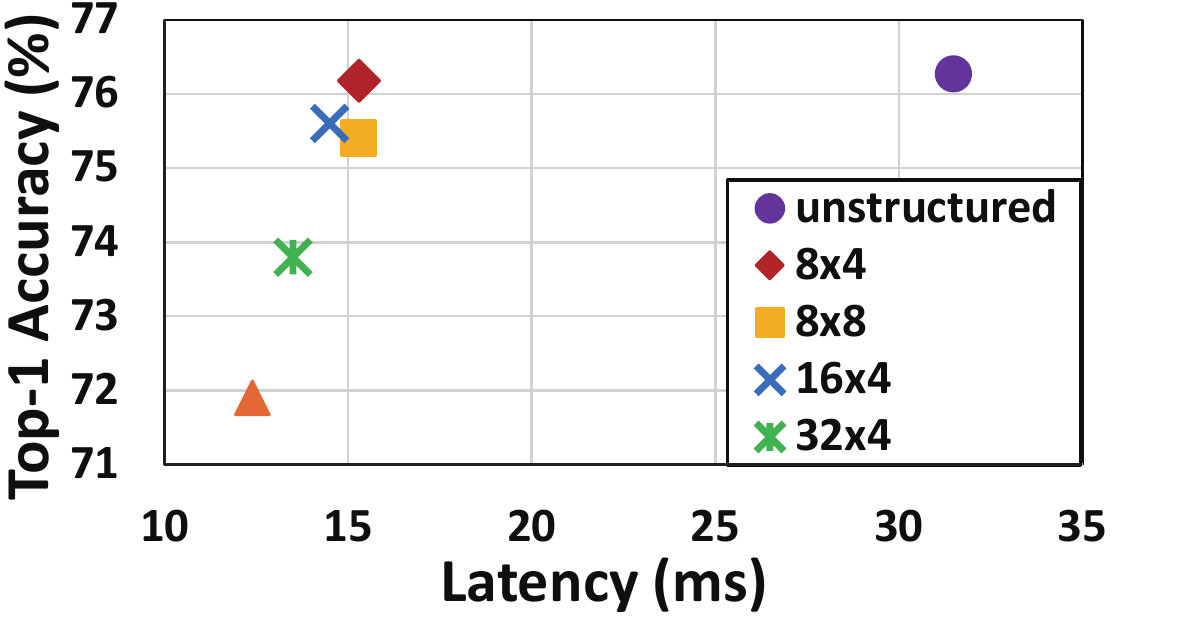}
    \caption{Accuracy vs. Latency (on a mobile phone) with different block sizes on ImageNet using ResNet-50 under uniform 6$\times$ pruning rate.}
    \label{fig:block_result}
\end{figure}

Figure~\ref{fig:block_result} shows example results of the accuracy vs. latency when applying block-based pruning on ResNet-50 (ImageNet dataset) with different block sizes.
A uniform pruning rate (i.e., 6$\times$) and block size are adopted through all layers.
Under the same pruning rate, non-structured pruning preserves the highest accuracy but has the worst performance in latency. On the contrary, coarse-grained structured pruning (i.e., the whole weight matrix as a block) achieves the lowest latency but with a severe accuracy degradation.
The results of block-based pruning show high accuracy and high inference speed (low latency) simultaneously. The reason is that the maximum hardware parallelism is limited by computation resources. Thus, even when dividing weights into blocks, each block's remaining weights are still sufficient to fulfill on-device hardware parallelism, especially on resource-limited mobile devices. 

\begin{figure}
    \centering
    \includegraphics[width=0.75 \textwidth]{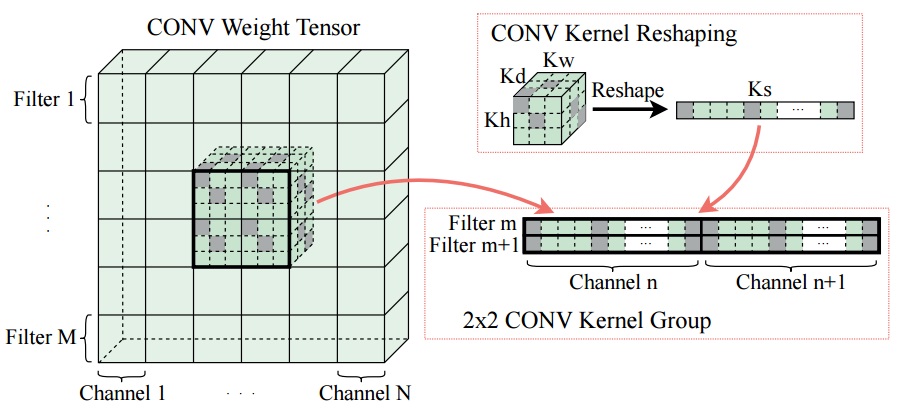}
    \caption{The generalized block-based pruning that applies to 3D convolutions.}
    \label{fig:3D_conv}
\end{figure}

The proposed block-based pruning results in effective acceleration while maintaining high accuracy on camera and LiDAR-based object detection \cite{cai2021yolobile,zhao2021neural} as well as RNN, transformer-based NLP applications \cite{li2020efficient,dong2020rtmobile}. Moreover, we have generalized block-based pruning to 3D convolutions \cite{niu2021rt3d} (Figure \ref{fig:3D_conv}), which apply to activity detection, 3D LiDAR-based detection, 3D reconstruction, and other sophisticated computer vision tasks. This is an important advantage of XGen. As an example of activity detection (details in \cite{niu2021rt3d}), we achieve over 20X speedup while maintaining accuracy compared with competing frameworks.

\subsection{High-Level Optimization}
\label{sec:highlevel}

As Figure~\ref{fig:xgen} shows, after XGen model optimizer reduces the size and complexity of the DNN model, there are a set of high-level optimizations, trying to optimize the computational graph structure, thus reducing the computation and intermediate result access and improving the computation parallelism. 

\begin{figure}[t]
    \centering
    \includegraphics[width=1 \textwidth]{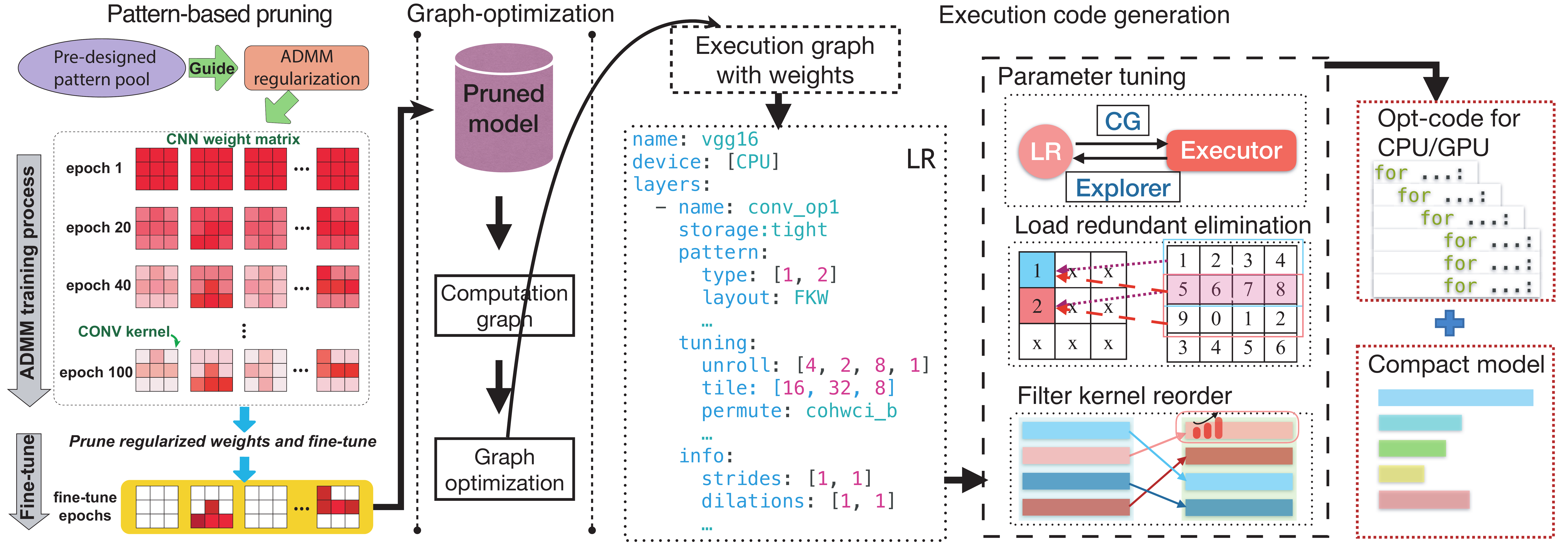}
    \caption{Overview of CoCoPIE's model and code optimization framework for DNN (using pattern-based pruning as an example).}
    \label{fig:system-overview}
\end{figure}

\subsubsection{Computational Graph Opt I: Graph Rewriting}

\begin{figure*}[t]
    \centering
    \includegraphics[width=0.96 \textwidth]{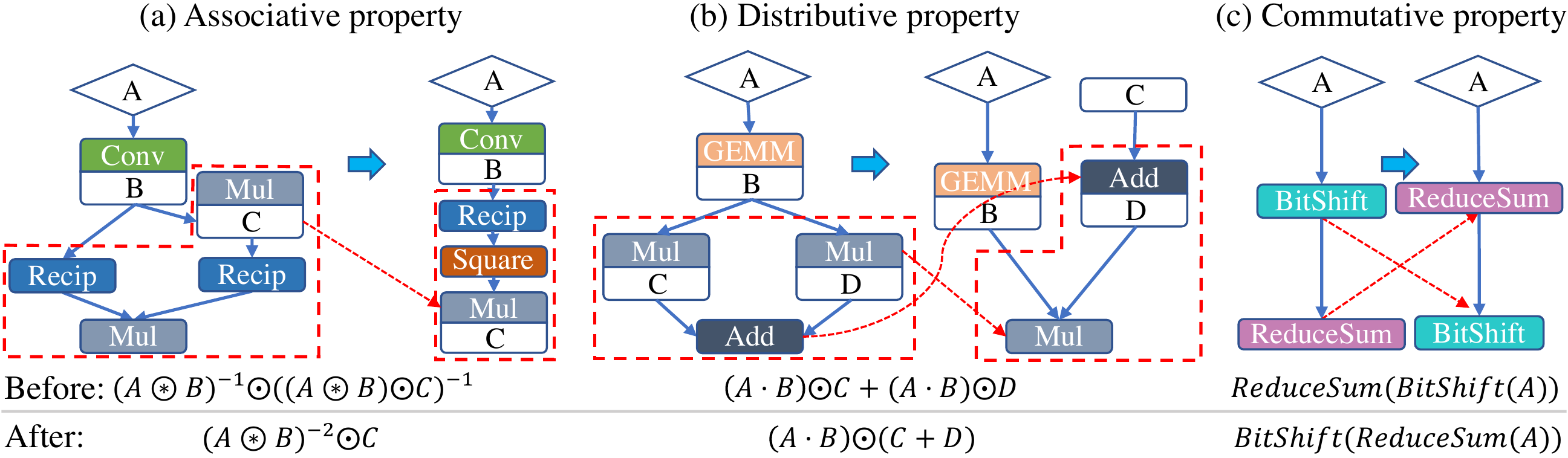}
    \caption{{\bf Examples of graph rewriting with mathematical properties.} Associative property
    explores the optimal execution order of operators and replaces the expensive combination of operators with a cheaper one.
    Distributive property
    explores the common combination of operators and simplifies the computation structure.
    Commutative property switches the execution order of operators to reduce the overall computation. 
    Note: the letter below each operator (e.g., {\tt B} below {\tt Conv} in (a)) or the letter in rectangle (e.g., {\tt C} in (b)) denotes that this input is from model weights rather than an intermediate result. The letter in diamond (e.g., {\tt A}) means that this is the input of this operator block, which could be the input of the model or intermediate result from a prior block. The intermediate results within this block are omitted for readability.}
    \label{fig:fusion-type-example}
\end{figure*}

The first high-level optimization of XGen is computational graph rewriting. It employs a novel mathematical-property based graph rewriting pass to optimize the computational graph. 
With this pass, XGen is able to 1) remove unnecessary operations, 2) eliminate redundant intermediate data copies, and 3) replace costly (combination of) operators  with more efficient ones.  
This graph rewriting carried out here  is in the spirit of the classical compiler optimization of strength reduction~\cite{cooper2001operator}; however, here it is  performed   on  complicated operators on  matrices or tensors rather than  on scalar expressions. 
Moreover, the rules we present are more complex and involved, and are based on operations that are common in DNNs. 
More importantly, compared  to existing efforts on computational graph substitution (e.g., TASO~\cite{jia2019taso}), our  graph rewriting is designed to work in conjunction with the subsequent high-level optimization (operator fusion) and identifies a set of operators and rules  for that specific purpose.  
Our evaluation results show that with graph rewriting, 
there are 18\% fewer fused layers left after fusion on GPT-2. 
Figure~\ref{fig:fusion-type-example} shows specific examples of  leveraged mathematical properties (distributive, communicative, and associative).

\subsubsection{Computational Graph Opt II: Universal DNN Operator Fusion} 

The second high-level optimization (on computational graphs) fuses DNN operators in a creative way. To achieve high accuracy, DNN models have become increasingly deep with hundreds or even thousands of operator layers (e.g., various transformers and the cutting-edge vision transformers). This trend causes two consequences: 
First, models with more layers usually generate more intermediate results, thus increasing the memory/cache pressure.
Second, deep models usually have an insufficient amount of
computations in each layer, thus degrading the processor’s
utilization, particularly for GPUs. 
Operator fusion (or kernel/layer fusion) can be an effective technique to reduce memory requirements and improve efficiency, and is a key optimization in many state-of-the-art DNN execution frameworks, such as TensorFlow, TVM, and MNN.
However, these frameworks usually adopt fusion approaches based on certain patterns that are too restrictive to cover the diversity of operators and layer connections. Polyhedral-based loop fusion techniques, on the other hand, work on a low-level view of the computation without operator-level information, and can also miss potential fusion opportunities. 


To address this challenge, 
the solution from CoCoPIE proposes a rigorous and extensive loop fusion framework called DNNFusion that can exploit the operator view of computations in DNNs, and yet enable a set of advanced transformations.  
The core idea is to classify operators into different types, and develop rules for different combinations of the types, as opposed 
to looking for patterns with specific combination of operations.  Particularly, DNNFusion first classifies the existing operations in a DNN into several groups based on the mapping relation between their input and output (such as One-to-One, One-to-Many, and others).   
Then,  DNNFusion leverages a mapping type analysis to infer  the profitability of different fusing combinations of these types of operators, binning the combination into 
three groups: likely profitable (and legal), likely not profitable, and ones where profitability may need to be determined through profile information. Table~\ref{tab:mapping-type-lookup} shows the details of this analysis.
%
The rest of DNNFusion framework comprises 
algorithms for determining fusion of specific operations (based on certain heuristics) and generating optimized fused code.  

\begin{table}
\caption{{\bf Mapping type analysis.} The first  column  and the first row (both without color)  show the mapping types of first and second  operators, respectively, 
before fusion, and the colored cells show the mapping type of the operator after fusion. Green implies that these fusion combinations can be fused directly (i.e., they are profitable). 
Red  implies that these fusions are unprofitable.   Yellow implies that  further profiling is required to determine profitability. }
\label{tab:mapping-type-lookup}
\centering
\includegraphics[width=0.6\columnwidth]{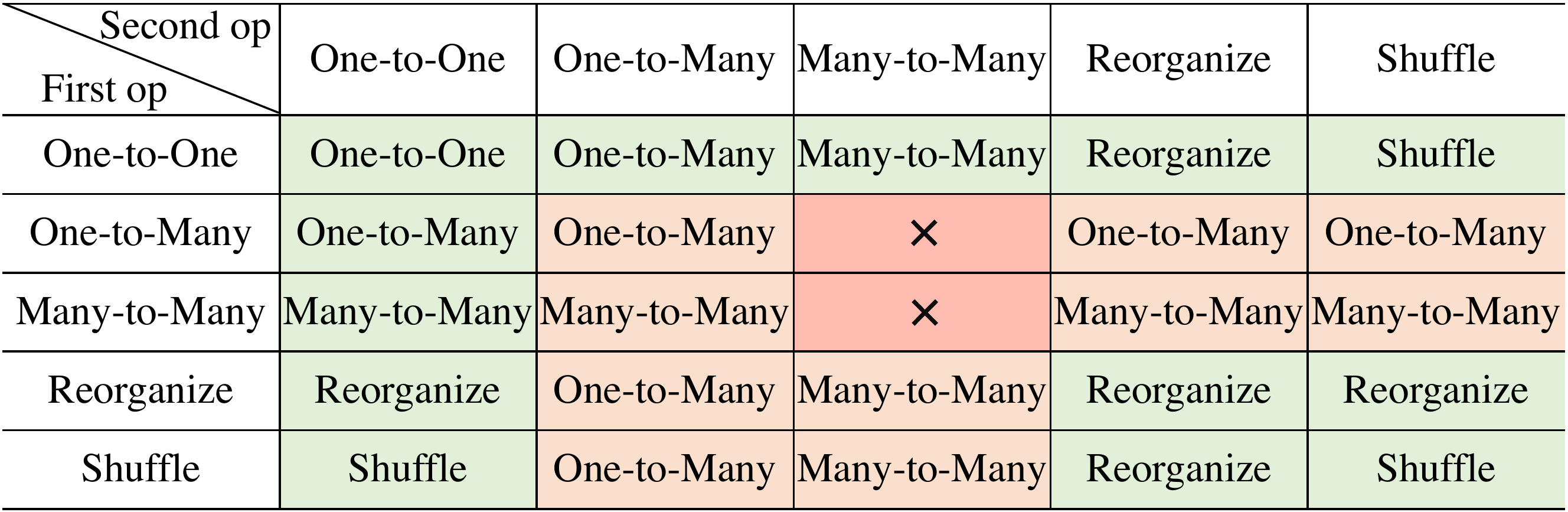}
\end{table}


DNNFusion has been extensively evaluated on a number of DNN models with varied types of tasks, model sizes, and layer counts. The results show that DNNFusion finds up to 8.8x higher fusion opportunities, outperforms four state-of-the-art DNN execution frameworks with 9.3x speedup. The memory requirement reduction and execution speedups enable many models on edge and mobile devices. Moreover, DNNFusion is especially effective in supporting extremely deep, next-generation NLP transformers (e.g., GPT) and {\em vision transformers}.

\subsection{Low-Level Optimization} 

After the high-level optimization streamlines the computations of the DNN, the low-level optimization ensures that efficient (parallel) code be generated for each layer in the DNN. The code should be able to fully capitalize the important hardware features and minimize the unnecessary computations for efficiency. Our optimizations at this level harness the opportunities from the regularity of the DNN sparsity created by the pattern- and 
block-based pruning at the model level. 

\subsubsection{Pattern-Conscious Code Generation}

\begin{figure}
    \centering
    \includegraphics[width=0.6 \textwidth]{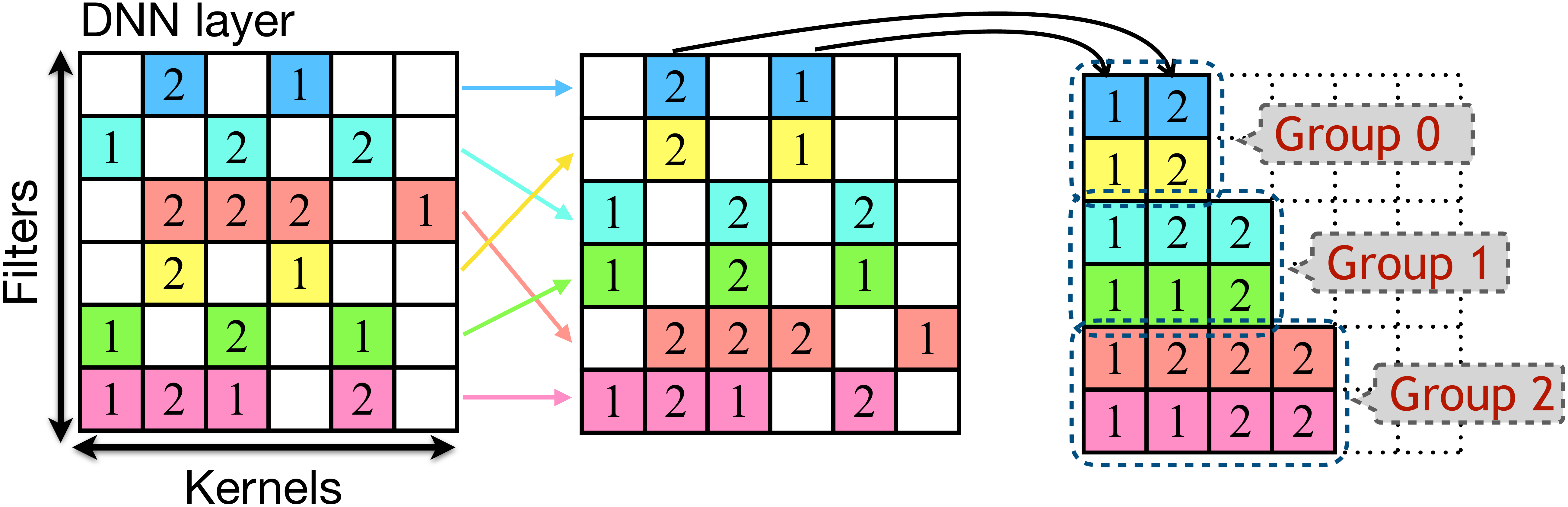}
    \caption{An example of filter kernel reorder. Each cell is a CONV kernel and the number on it specifies its pruning pattern style. After filter kernel reorder, each group will be executed concurrently by multiple threads with each thread processing a filter. This design improves critical performance concerns like heavy control-flow instructions, thread divergence and load imbalance.}
    \label{fig:kernel-reorder}
\end{figure}

Another key low-level optimization is to support pattern-conscious code generation through a set of compiler and parallel computing techniques (as shown on the right half of Figure~\ref{fig:system-overview}).  
This code generation is based on a high-level fine-grained Layerwise Representation (LR) that captures the DNN sparsity information. This LR also includes intensive DNN layer specific information to enable aggressive layerwise optimizations. In particular, it includes detailed kernel pattern and connectivity-related information (e.g., the pattern types presented in this layer, the pattern order in each filter, the connection between kernels and input/output channels, etc.); and tuning-decided parameters (e.g., the input and output tile sizes, unrolling factors, the loop permutation of this layer, etc.). 
We particularly emphasize two critical optimizations in this code generation:

\noindent{\bf Filter Kernel Reorder and Compact Filter-Kernel-Weight (FKW) Storage: } It proposes filter kernel reordering to address two key challenges: heavy control-flow instructions, and thread divergence and load imbalance. The insight is that for a specific DNN layer, the patterns of all kernels are already known after model training, so the inference computation pattern is also known before model deployment. Kernel reordering leverages this knowledge to organize the filters with similar kernels together to improve inter-thread parallelization and order the same kernels in a filter together to improve intra-thread parallelization. Figure~\ref{fig:kernel-reorder} illustrates the basic idea of kernel reordering.
After kernel reordering, our LR stores the DNN’s weights in a novel compact Filter-Kernel-Weight format (called FKW \cite{niu2020patdnn}). Compared with existing compact data formats (like CSR), FKW is higher-level and results in much less extra structure overhead (i.e., the total size of all index arrays that are used for weights data access). In addition, FKW leverages the pattern information, and stores the kernels with the kernel reordering information that will support later branch-less DNN execution. Other compact data format cannot support this.
    
\noindent{\bf Load Redundancy Elimination:} We propose two techniques to improve the memory performance and eliminate data load redundancy caused by irregular memory access (in the form of array indirection): an effective input tiling to improve the cache performance, and the optimized code generation that eliminates redundant memory loads with the help of the pre-defined pattern information. The second one is particularly interesting.
Our key insight for this load redundancy elimination is: in DNN execution, such as a convolution operation, the data access pattern of the input and output is decided by the (none-zero elements) patterns of kernels that are already known after training. Therefore, it is possible to generate the optimized data access code with this information for each pattern of kernels and call them dynamically during the DNN execution. The generated codes consist of all statically determined data access instructions for the kernel-level computation with a careful instruction reorganization to 1) eliminate all indirect memory accesses; and 2) eliminate all redundant {\em register} load operations.

\subsubsection{Deep Reuse}

Besides the low-level code generation, another technique {\em deep reuse} offers a method from the dimension of activation maps to further reduce the amount of unnecessary computations. It speeds up DNN training and inference through discovering and exploiting deep reusable computations on the fly. It is effective, halving the inference time of CNNs implemented on state-of-the-art high performance libraries and compression techniques, while causing virtually no ($<$0.0005) accuracy loss. It is meanwhile easy to use, requiring no special hardware support or CNN model changes, ready to be applied on today's systems. 

\begin{figure}
	\begin{center}
		\includegraphics[width=.7\columnwidth]{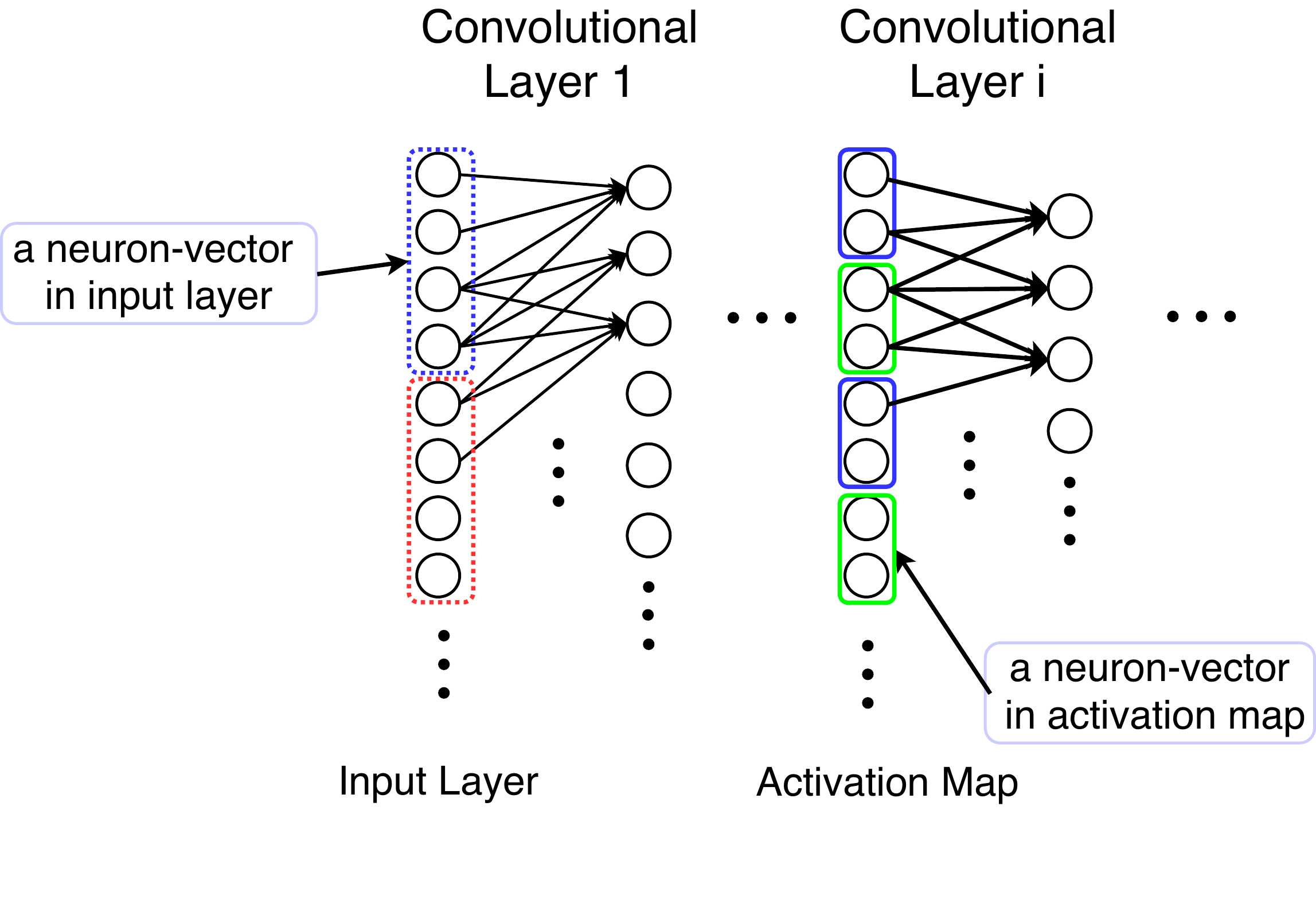} 
		\caption{Illustration of a simple 1-D CNN. The input for convolutional layer 1 is called the input layer while the input for convolutional layer i with $i\neq 1$ is called the activation map. Neurons in the same block form a neuron-vector. Block colors indicate the similarity of the neuron-vector values.}
		\label{fig:all_terms}
	\end{center}
\end{figure}

{\em Deep reuse} centers around similarities among neuron vectors. 
A {\em neuron vector} is made up of values carried by some consecutive neurons at a CNN layer. As Figure~\ref{fig:all_terms} illustrates, if the layer is an input image layer, a neuron vector contains the values of a segment of input image pixels; if the layer is a hidden layer, it contains a segment in its activation map.

\begin{figure}
	\begin{center}
		\includegraphics[width=0.8\columnwidth]{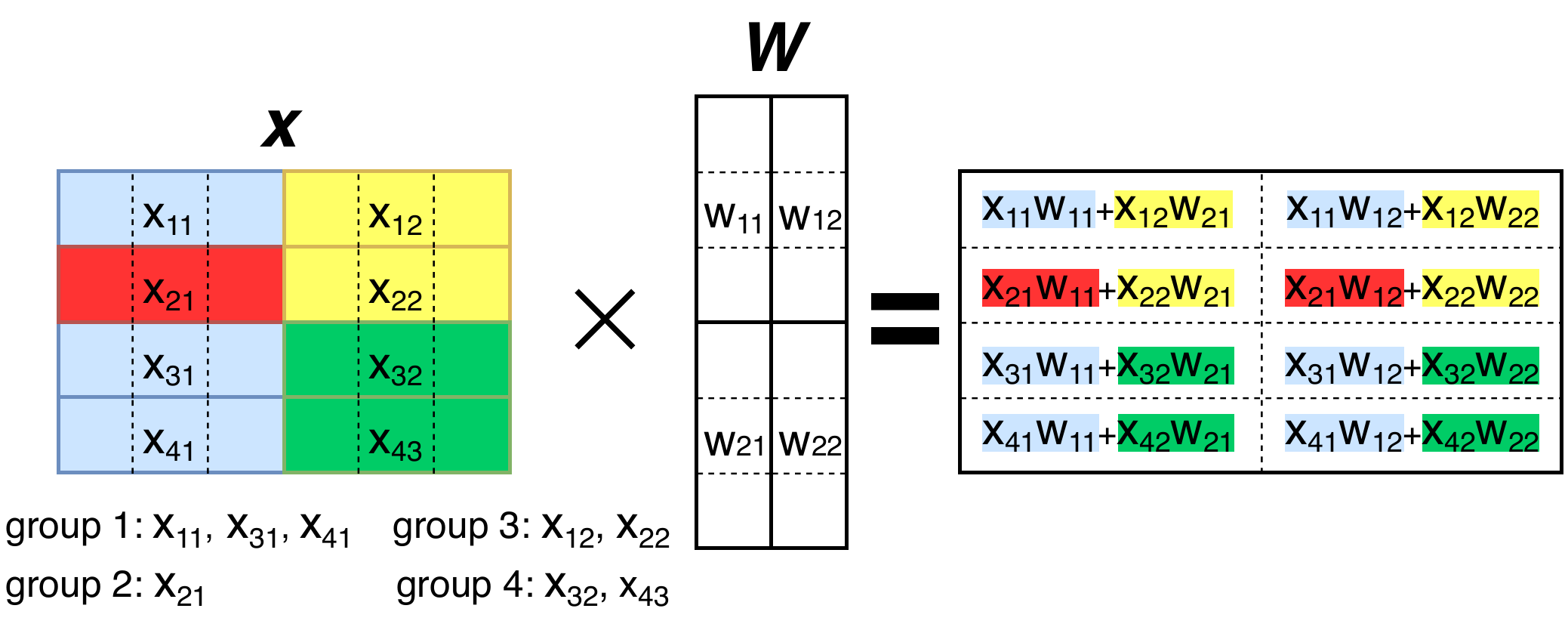} 
		\caption{An example of the basic form of computation reuse across neuron vectors in convolution $X\times W$. Instead of calculating 16 dot products, we only need to compute 8 of them: $\Vec{x}_{11} \cdot \Vec{w}_{11}$, $\Vec{x}_{11} \cdot \Vec{w}_{12}$, $\Vec{x}_{21} \cdot \Vec{w}_{11}$, $\Vec{x}_{21} \cdot \Vec{w}_{12}$, $\Vec{x}_{12} \cdot \Vec{w}_{21}$, $\Vec{x}_{12} \cdot \Vec{w}_{22}$, $\Vec{x}_{32} \cdot \Vec{w}_{21}$and $\Vec{x}_{32} \cdot \Vec{w}_{22}$.}
		\label{fig:basicReuse}
	\end{center}
\end{figure}

The basic idea of {\em deep reuse} is to leverage similarities among neuron vectors, such that computation results attained on one neuron vector can be effectively reused for some other neuron vectors in CNN inferences. Figure~\ref{fig:basicReuse} illustrates the basic form of such reuses. The eight 3-neuron vectors, represented by $\Vec{x_{ij}}$, form four groups. Neuron vectors in a group are similar to each other. In this example, when the dot product of one of them is reused for all others in the group (e.g., $\Vec{x_{11}}\cdot \Vec{w_{11}}$ for $\Vec{x_{31}}\cdot \Vec{w_{11}}$ and $\Vec{x_{41}}\cdot \Vec{w_{11}}$), half of the computations in $X\times W$ could be saved.

Deep reuse can be implemented through Locality Sensitive Hashing (LSH), an online data clustering method. The computation reuse can have multiple levels, within an input item, within a batch of inputs, or across batches. By effectively exploiting the redundancy in inputs and activation maps, it reduces DNN computations significantly and brings substantial performance benefits. More details are given in our papers~\cite{Ning+:ICDE19,Ning+:ICS19}.

\subsection{Compiler-Aware Neural Architecture \& Pruning Co-Search (CAPS)}
\label{sec:naps}



\begin{figure}
    \centering
        \includegraphics[width=0.7 \textwidth]{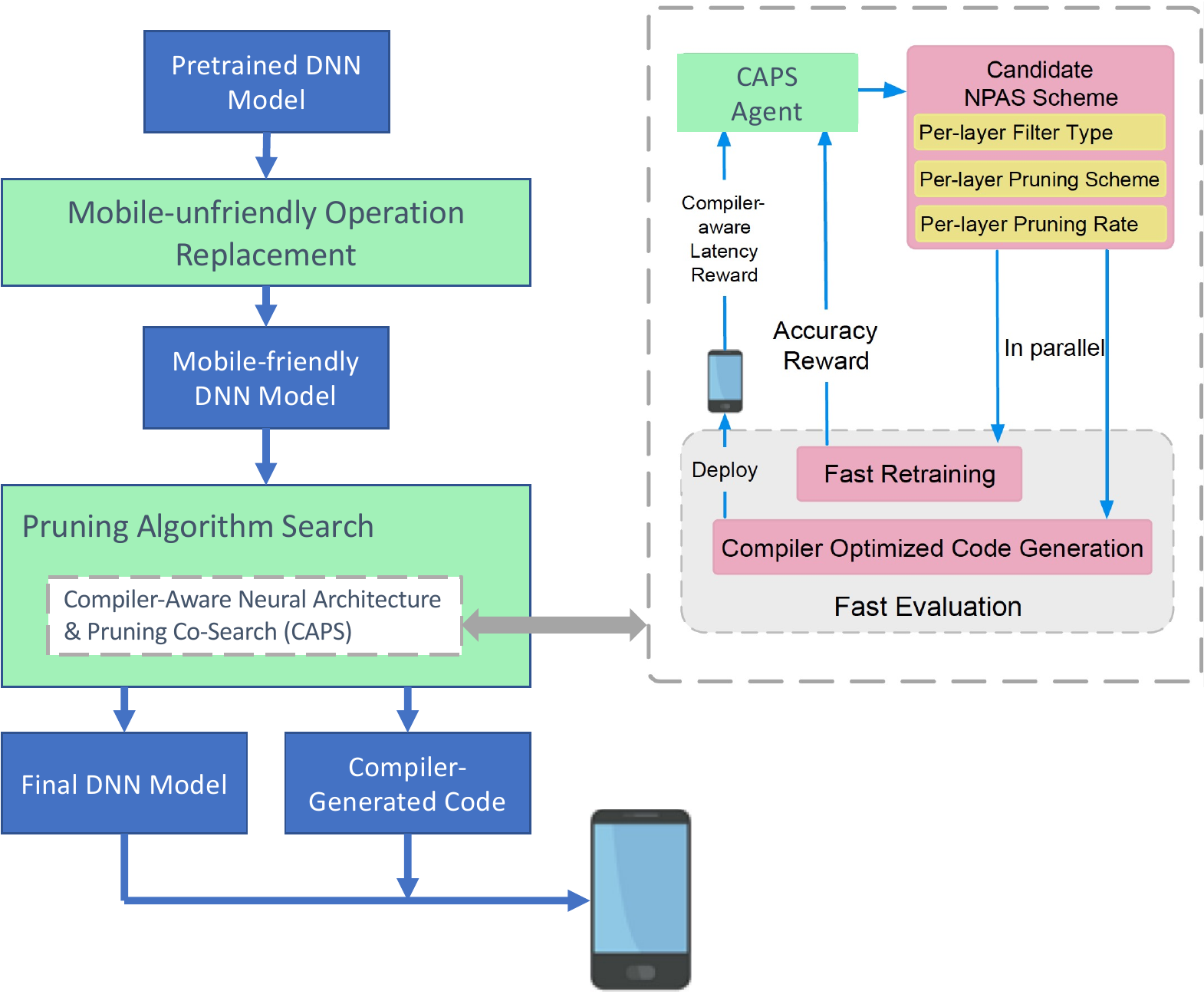}
    \caption{Overview of the proposed CAPS framework, which simultaneously optimizes neural network architecture and pruning schemes.}
    \label{fig:npas}
\end{figure}
\begin{figure}
    \centering
    \includegraphics[width=0.65 \textwidth]{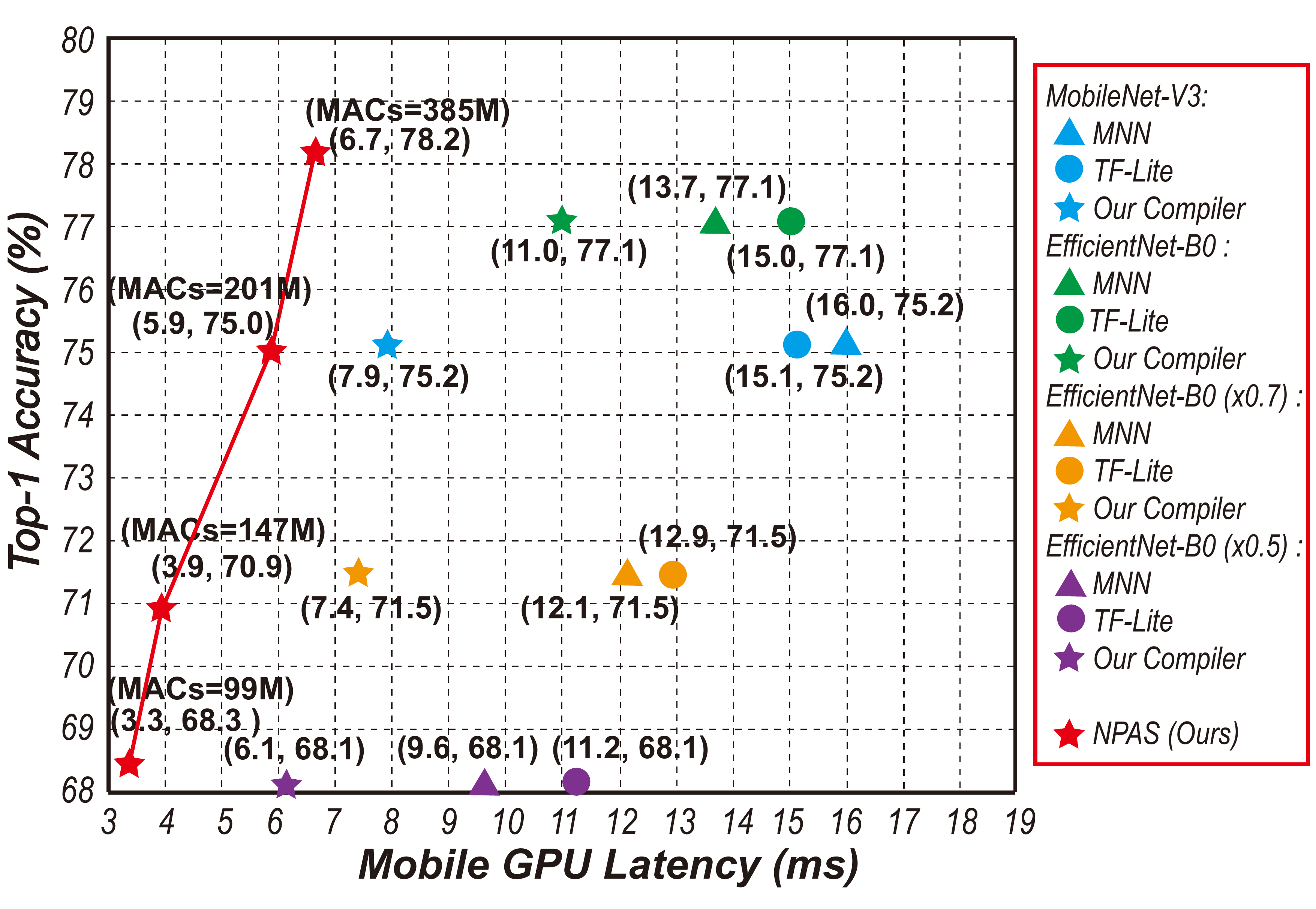}
    \caption{Top-1 Accuracy vs. Latency comparison on mobile GPU (on Samsung Galaxy S10 mobile phone) on ImageNet dataset.}
    \label{fig:mobilegpu_npas}
\end{figure}

The three main components of XGen is tied together with CAPS, which iteratively identifies the best model and code optimization parameters as well as model architectures. 

While our compiler optimizations provide notable mobile acceleration and support various sparsity schemes, it introduces \emph{a much larger model optimization space}. 

An active research area is the Neural Architecture Search (NAS), which designs more efficient DNN architectures using automatic searching algorithms. Some recent work acknowledge the importance of hardware-software co-design and incorporate the inference latency into NAS, which can achieve better result than the intuitive volume estimation using the number of weight parameters or computations. However, none of these hardware-targeting work fully exploit the potential of compiler optimizations or satisfy an overall latency requirement. Moreover, the steps of model compression, compiler optimization and network architecture search (NAS) are largely performed independently in prior work. 

It is desirable to perform a \emph{joint network pruning and architecture search with compiler-based code optimizations included in the loop}, determining the best filter type and size, as well as pruning scheme and rate, for each individual layer. CAPS is designed to that end. It can be configured to meet various objectives, such as to maximize accuracy while at the same time satisfying the DNN latency constraint on the target mobile device. 

The overall workflow of CAPS~\cite{li2021npas} is shown on the left side in Figure~\ref{fig:npas}. It first prepares the input DNN model by replacing some mobile-unfriendly operations with friendly ones. It then enters the main loop. At the outermost level is the trials of different pruning algorithms, which determine what search algorithms to use for the co-search of the neural architecture and pruning. For a given pruning algorithm, NAPS effectively explores the neural architecture and pruning space in a hand-in-hand manner to find the best pruned model. NAPS includes code-generation and performance assessment in the loop to ensure the speed of the generated DNN model. As the process exhibits a larger search space than prior NAS work, to perform efficient search, NAPS employs a meta-modeling procedure based on reinforcement learning (RL) with fast evaluation and Bayesian optimization. It makes the total number of training epochs comparable with that in the state-of-the-art NAS frameworks. 

To further reduce the search in the huge pruning space, XGen explores {\em composability}, a property (fundamental in software engineering) that we discovered in the training of a collection of pruned CNN models. The basic observation is that two candidate CNN networks in the pruning space often differ in only some layers, and the training results of the common layers can be reused across networks to save some training time. More generally, our solution views the networks to search as compositions of a set of building blocks (a {\em block} is a sequence of CNN layers). It pre-trains (some of) these building blocks and then assembles them into the to-be-explored networks. 

To identify the best set of building blocks to pre-train and maximize the benefits, it uses a novel algorithm, which represents all layers of all to-be-explored networks as a sequence of symbols, and uses a hierarchical compression algorithm Sequitur~\cite{nevill1997identifying} to produce a context free grammar (CFG) and uses it to quickly find out the most reusable building blocks~\cite{Guan+:PLDI2019}.



The NPAS framework in XGen achieves by far the best mobile acceleration results (shown in Figure \ref{fig:mobilegpu_npas}). For example, the inference times with ImageNet are 6.7ms, 5.9ms, and 3.9ms with 78.2\%, 75\%, and 71\% Top-1 accuracy, respectively, on an off-the-shelf mobile phone. The unique features of XGen include the following:

\subsection{XGen Runtime}

The first three components of XGen output highly optimized DNN code. {\em XGen runtime} helps ensure that the DNN code can actually achieve a high speed in its executions, regardless of the hardware differences or interferences from other applications on the same device. XGen runtime is not mandatory for the optimization result of XGen to be used, but could make large differences in an environment that requires dynamic adaptations, due to either hardware diversity or serious resource contention. 

In our design, XGen runtime has two main functionalities: (i) helps the DNN code from XGen adapt its internal paths or components (functions, instructions, data layouts) to fit the underlying hardware, especially when the target hardware is diverse (e.g., different kind of smartphones of the users of a popular app); (ii) coordinates the usage of computing resources (e.g., computing units, memory) among co-running DNNs. 

For the first functionality, XGen has a design of {\em plastic intermediate representation (IR)} for representing DNN code, which can be leveraged by XGen runtime to adapt DNN code on the fly to best fit the underlying hardware. 

For the second functionality, XGen has a design of {\em AI-conscious DNN co-scheduling}, which takes DNN properties and (offline and online) DNN optimizations into consideration when coordinating resource usage. The environments can be closed where the set of co-running tasks are predefined, or open where the set of co-running tasks dynamically change and are hard to predict. 
XGen runtime has designs for both closed and open environments.

\begin{figure}
    \centering
    \includegraphics[width=.6\textwidth]{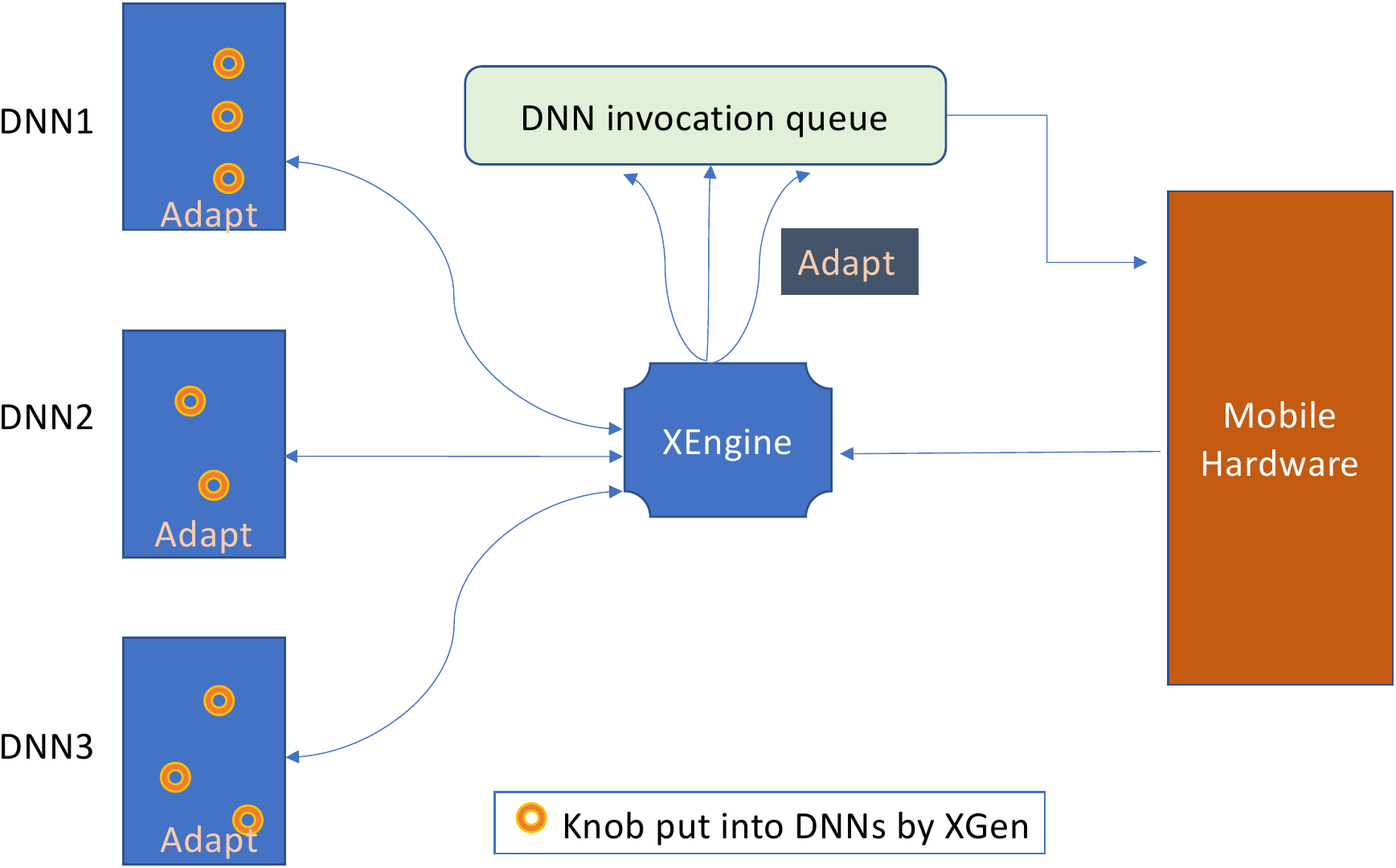}
    \caption{The adaptations of DNN executions on a device by XEngine are through knobs injected into DNNs by XGen as well as the scheduling by XEngine runtime.}
    \label{fig:xengine}
\end{figure}

These two functionalities are coupled, served together by XGen runtime. As Figure~\ref{fig:xengine} shows, the adaptations are enabled via knobs injected into DNNs by XGen as well as the scheduling by XEngine. We call it {\em synergistic adaptation}. An example is the use of GPU on a smartphone. The knobs inside a DNN---such as at which layer to exit on a multi-exits DNN---may allow the adjustment of the amount of computations the DNNs may impose on a GPU, while the scheduling by XEngine may allow the DNN to time-share the GPU with other DNNs on the device in a desirable manner. 

\begin{figure}
    \centering
    \includegraphics[width=0.7\textwidth]{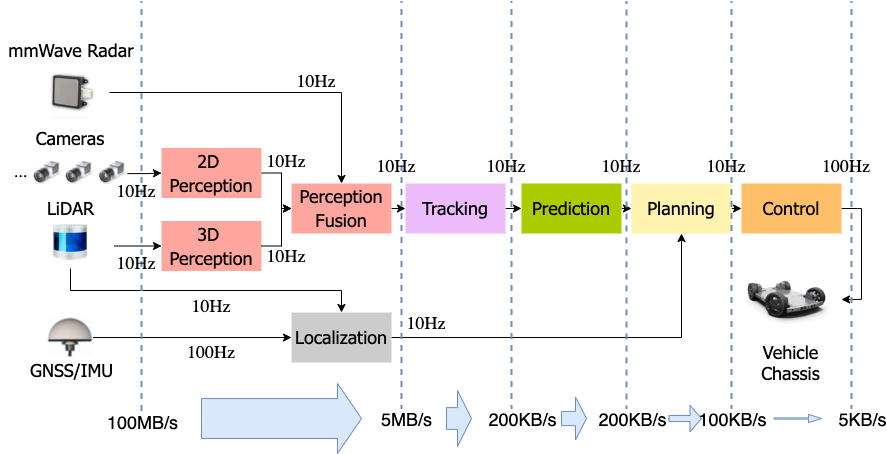}
    \caption{High-level workflow of a level-4 autonomous vehicle. The input of each camera and lidar device is fed into a DNN-based 2D or 3D perception module.}\label{fig:am}
\end{figure}

We next draw on autonomous vehicle as an example closed environment to briefly showcase the benefits from XEngine runtime. An autonomous driving application consists of a predefined set of task modules that run on the system concurrently. These task modules involve complicated dependences, with DNN playing an important role in the modules as illustrated in Figure~\ref{fig:am}. 

Even though there are runtime systems dedicated to autonomous driving, they work poorly in resource-constrained scenarios. The realtime runtime in the industry, ROSCH, for instance fails delivering real-time responses when it runs L4 autonomous driving applications on an NVIDIA Xavier Jetson card (Section~\ref{sec:quantitative}). 

XGen runtime features {\em DNN-oriented heterogeneous scheduling}. The scheduling addresses limitations of existing scheduling algorithms used in both real-time systems and conventional operating systems, such that when multiple DNNs execute on a device at the same time, they can all run efficiently by taking a full advantage of the processing units and memory on the device. 

More specifically, the design addresses three major limitations of existing runtime systems. 

\begin{itemize} 
\item Limitation I: Starvation happens when prior scheduling schemes are applied to applications with multiple DNNs deployed on a single resource-constrained device. 

Our solution features {\em just-in-time priority adjustment}, which resolves the starvation by adjusting the affinity and priorities of tasks in a just-in-time manner. 

\item Limitation II: Some types of accelerators are left substantially under-utilized due to the hardware-oblivious model designs and implementations.

Our solution employs {\em model-schedule co-optimization}, an approach that customizes the optimization of DNN models based on the constraints of the underlying hardware to maximize the hardware utilization. It further integrates DNN scheduling into the model optimization process to iteratively determine the best model optimizations and the corresponding schedules in a hand-in-hand manner to achieve the desirable accuracy-speed tradeoff.

\item Limitation III: Current scheduling algorithms cannot deal with hybrid workloads that can employ multiple types of accelerators.

Our solution offers {\em DAG-instantiating scheduling}, an approach that extends the current scheduling algorithms to better manage the heterogeneous computing resources. To accommodate the different performance of a DNN model on different types of computing units, it creates an algorithm to efficiently enumerate the combinations of DNN models on all types of computing units according to their dependence, based on which, it produces the schedules that best tap into the full potential of all types of computing units for the inter-dependent DNN tasks.
\end{itemize}

The proposed runtime techniques prove effective. When being applied to autonomous driving workloads, it makes complex workloads able to achieve realtime performance on a low-end device (Xavier Jetson), which costs 10X less than the devices that the industry currently requires for those workloads as Section~\ref{sec:quantitative} shows. For details, please refer to our paper~\cite{sung2021enabling}.

\vspace{.3in}

These four core technologies respectively focus on four different levels of the DNN software stack: Model, computation graph, code generation, and runtime. But at the same time, they form a synergy: The model optimizations and the graph and code optimizations are designed hand in hand, and they all help reduce the computing and memory demands of DNNs, paving the path for the runtime techniques to exert its full power. Such a full-stack coordinated optimization distinguishes CoCoPIE solutions from any other existing solutions. It is the reason that CoCoPIE can achieve multiple times of better results than other solutions. 

\section{Comparisons}
\label{sec:compare}

There are many other DNN optimizing frameworks and techniques developed in the recent several years. This section first lists the principle features of XGen and its qualitative comparison with several representative frameworks, and then provides quantitative comparisons with the state-of-the-art frameworks on the market on some common DNN models. 

\subsection{Qualitative Comparisons}

The principle features that distinguish XGen from other frameworks are three:
\begin{itemize}
    \item {\em Compression-compilation co-design:} This feature produces the unique {\em fine-grained pruning patterns} inside the coarse-grained structures, and the correspondingly tailored optimizations in the compiler. 
    \item {\em AI-aware co-optimizing runtime:} This feature in the XGen runtime produces the unique {\em model-schedule co-optimization} and the other just-in-time optimizations. 
    
    \item {\em Full-stack synergy:} XGen is the only DNN framework that includes optimizations from DNN models to code and runtime schedules, all designed in a hand-in-hand manner to form a coherent synergy. 
\end{itemize}

Table~\ref{tab:qualitative} summarizes the qualitative differences between XGen and some best-known competitors. Fundamentally, none of the previous frameworks fully share the three key features of XGen. Also OctoML (TVM) is less optimized for edge devices compared with servers. Furthermore, XGen supports next-generation DNN models such as extremely deep transformers and {\em vision transformers}.

\begin{table}
    \centering
    \caption{Qualitative Comparison between XGen and Competitors}
    \label{tab:qualitative}
    \begin{tabular}{|p{.3\textwidth}|p{.6\textwidth}|}\hline
    {\bf Competitors} & {\bf Differences} \\ \hline\hline
    PyTorch Mobile \cite{pytorchmobile}, TF-Lite \cite{tensorflowlite}, MNN \cite{ali-mnn}, SNPE \cite{snpe} & Siloed design in compression and/or compilation; no runtime scheduling; partial stack \\ \hline
    OctoML \cite{octoml} & Compilation only; no compression or runtime scheduling; no co-design \\ \hline
    Deci \cite{deci}, Neural Magic \cite{neuralmagic}, DeepLite \cite{deeplite} & Compression plus existing compilers, designed and developed separately; no runtime scheduling; partial stack \\ \hline 
    \end{tabular}
\end{table}

\subsection{Quantitative Comparisons} 
\label{sec:quantitative}

\subsubsection{Comparison Results on Off-the-Shelf Mobile Phone}


\begin{table}
    \centering
        \caption{Mobile comparison results on Samsung Galaxy S10 phone over different DNNs/applications under the same accuracy.}
    \label{fig:mobile_comparison}
    \includegraphics[width= \textwidth]{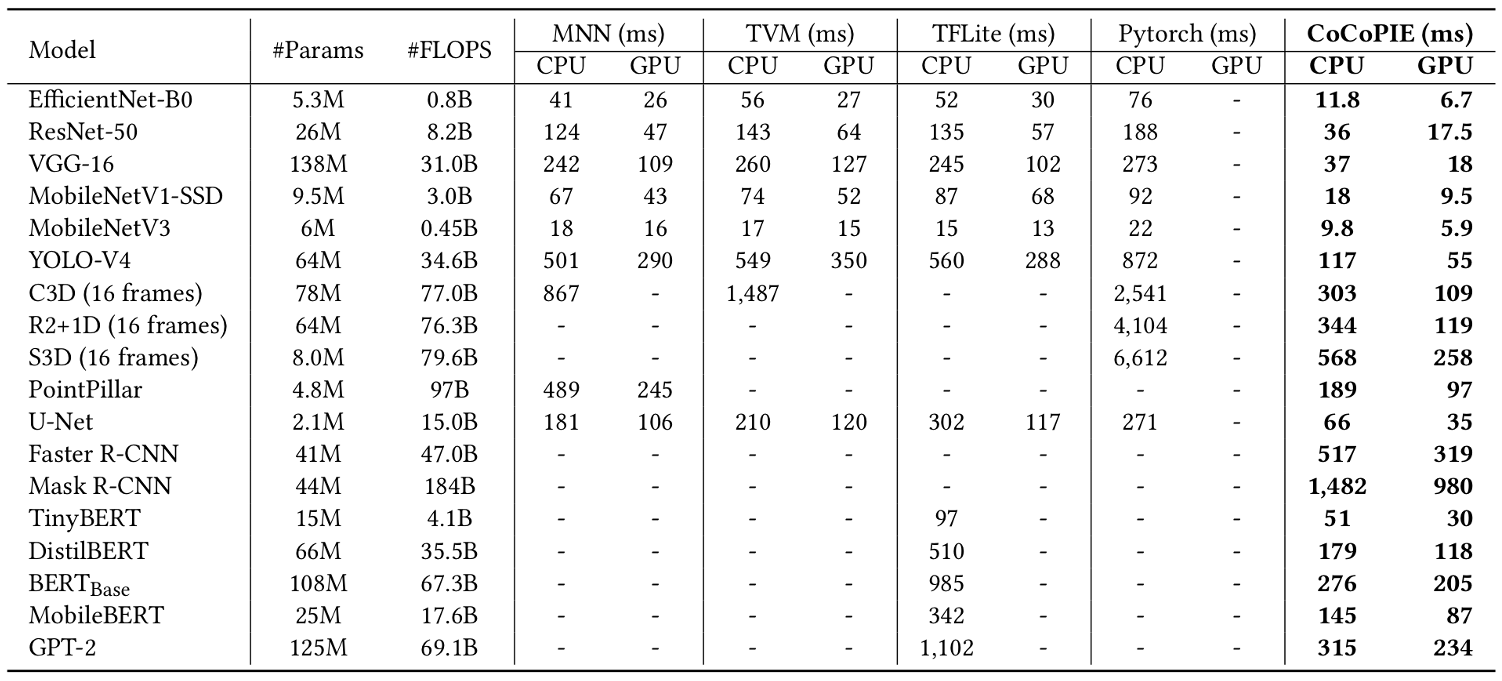}
\end{table}

\begin{figure}
    \centering
    \includegraphics[width= 0.56\textwidth]{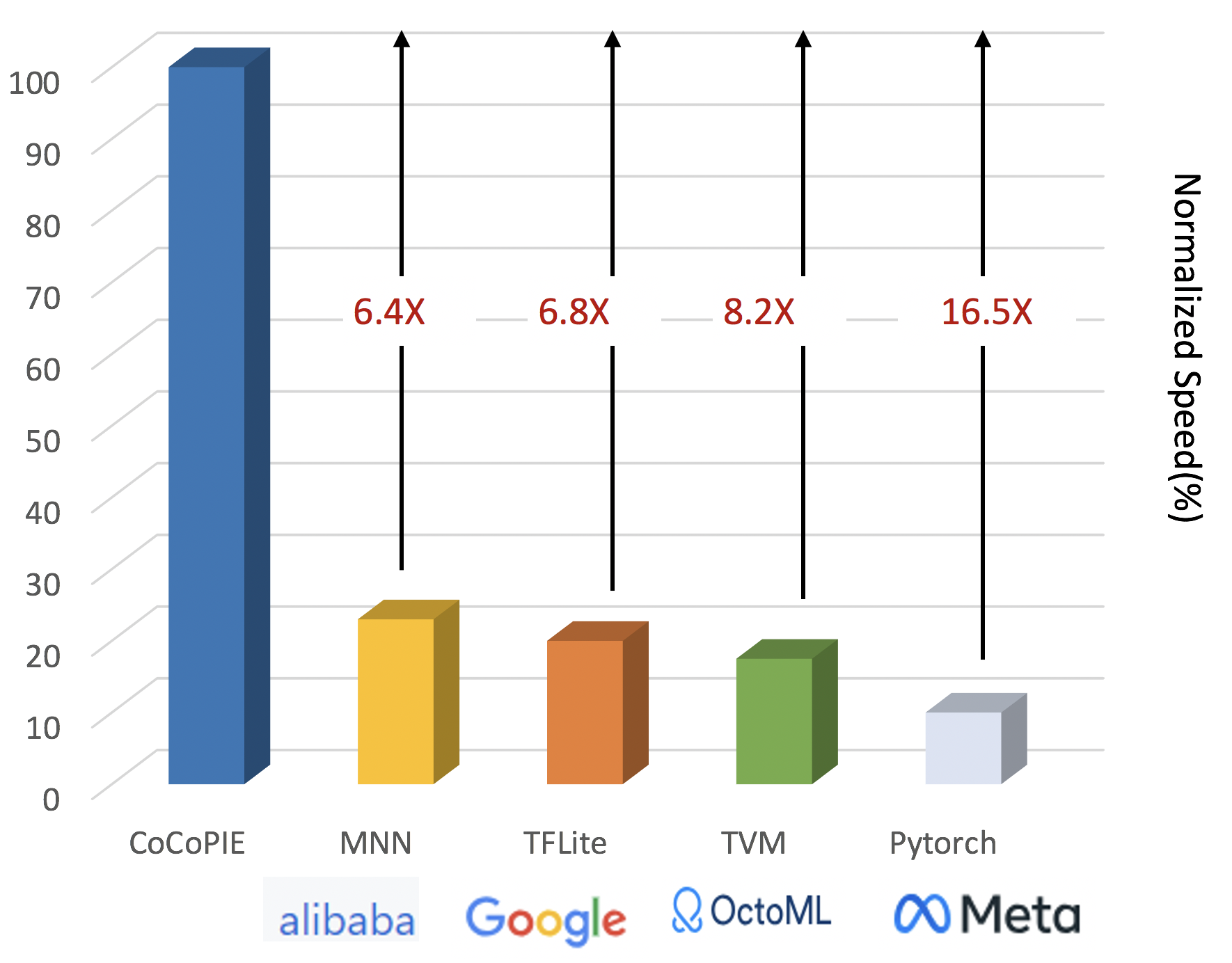}
    \caption{Average speedup summary on Samsung Galaxy S10 phone compared with the other frameworks under the same accuracy.}
    \label{fig:comparison_summary}
\end{figure}

Benefited from Compression-Compilation Co-Design, we evaluate \projectname on a Samsung Galaxy S10 cell phone with the latest Qualcomm Snapdragon 855 mobile platform that consists of a Qualcomm Kryo 485 Octa-core CPU and a Qualcomm Adreno 640 GPU. We perform a comprehensive evaluation on four categories of applications, image classification, object detection, semantic/instance segmenation, and transformer-based NLP applications, with representative, state-of-the-art DNN models. We compare with four state-of-the-art software acceleration frameworks TFLite~\cite{tensorflowlite}, TVM~\cite{chen2018tvm}, MNN~\cite{ali-mnn}, and PyTorch Mobile \cite{pytorchmobile}, \textbf{under the same testing accuracy} for these models. Table~\ref{fig:mobile_comparison} shows the detailed comparison results on mobile CPU and GPU, while Figure \ref{fig:comparison_summary} provides the summary. \projectname outperforms all other frameworks for all cases, and largely satisfies the overall real-time requirement on off-the-shelf mobile device. On average, XGen achieves 6.8$\times$, 8.2$\times$, 6.4$\times$, and 16.5$\times$ speedups on TFLite, TVM, MNN, and PyTorch Mobile, respectively, illustrating the effectiveness of compression-compilation co-design. Moreover, one significant advantage of XGen is on NLP and the recent advances of vision transformers, which are lack of support by other frameworks. Please see video demos at the CoCoPIE YouTube channel\footnote{\url{ http://www.youtube.com/channel/UCCKVDtg2eheRTEuqIJ5cD8A/}} and Bilibili channel\footnote{\url{https://space.bilibili.com/573588276?from=search&seid=11881710196887435131}}.

When using compiler only (i.e., comparing on the same model without compression or NAS), \projectname also consistently outperforms the other software acceleration frameworks, TFLite, TVM, MNN, and PyTorch Mobile by at least 2.5X speedup on average. This illustrates the significant advantages of our high-level and low-level compiler code generation optimizations. Please refer to \cite{niu+:PLDI21} for more details.

\textbf{Energy Efficiency Comparison:} \rv{In terms of energy consumption, \projectname is $8.0\times$ less than TVM. The power consumption rate of the entire mobile device is about the same as that of TVM executions, 3.8W (tested by Qualcomm Trepn power profiler). But its $8.2\times$ less execution time leads to the large savings in energy.}

\begin{figure}
    \centering
    \includegraphics[width= 0.8\textwidth]{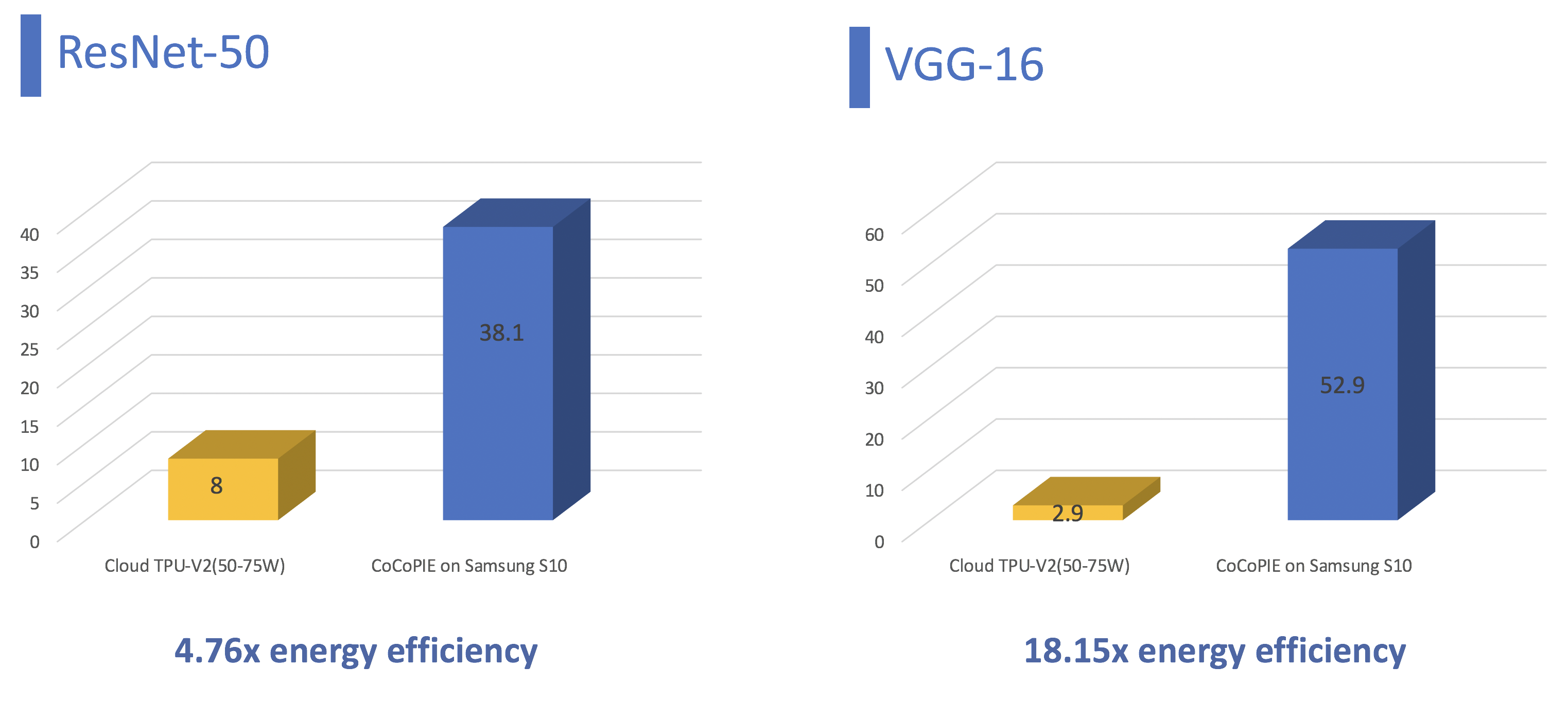}
    \caption{Energy efficiency comparison of CoCoPIE solution with ASIC Google's cloud TPU-V2.}
    \label{fig:energy_efficiency}
\end{figure}

The results also consistently outperform a number of ASIC and FPGA solutions in both performance and energy efficiency. Figure \ref{fig:energy_efficiency} demonstrates the comparison results on performance and energy efficiency of the CoCoPIE solution on off-the-shelf mobile device with special ASIC hardware including Google's cloud TPU-V2~\cite{googletpu}, on the same network models. Similar advantages of the CoCoPIE software solution can be found comparing with edge TPU~\cite{googletpu}, NVIDIA Jetson AGX Xavier, Cambricon MLU-100, Eyeriss~\cite{isscc_2016_chen_eyeriss}, etc., and a series of FPGA solutions in terms of accuracy, performance, and energy efficiency. Please refer to \cite{cocopieACM,yuan2021work} for more details. 

The better results of XGen come from three reasons: (i) the compression-compilation co-design more effectively matches models with hardware; (ii) smartphone chips are built with the most advanced technology (e.g., 7nm, 11nm technology), while FPGA/ASIC solutions are based on older and less energy-efficient 28nm or 40nm technologies. 
(iii) current ASIC/FPGA solutions are often optimized for a specific DNN type/size (e.g., edge TPU for small-scale DNNs, Cambricon MLU-100 for large-scale DNNs), while XGen, as a software method, can better adapt to all kinds of networks. 

\textbf{Comparison with NeuroMagic:} NeuroMagic focuses on generating sparsity (non-structured sparsity) on sample neural networks (ResNet-50 and YOLO-v3), and has a sparsity-aware inference engine on desktop CPU. Our \projectname outperforms NeuroMagic in (1) advanced sparsity and NAS schemes instead of the most inefficient non-structured sparsity, (2) advanced compiler-level code generation techniques, and (3) compression-compilation co-design. For a comparison under the same accuracy, NeuroMagic achieves 27ms inference time on MobileNet-V2 on an Intel 4-core CPU (power consumption $>$30W), while we achieve 3.3ms inference time on a 3.8W mobile platform. We achieve an energy efficiency gain of 64.6$\times$. For YOLO-based object detection, NeuroMagic achieves 36.2ms inference time on a 24-core Intel CPU (power consumption $>$100W), while we achieve 55ms inference time with the same accuracy on a 3.8W mobile platform. We achieve an energy efficiency gain of 17.3$\times$. The NeuroMagic results are from their webpage. Needless to say, our CoCoPIE solution also has much broader applications (including the most advanced, state-of-the-art DNN models), supporting platforms, and degree of automation.

\subsubsection{Comparison Results on Mobile DSP and MCU}

\begingroup
\setlength{\tabcolsep}{2.8pt}
\renewcommand\arraystretch{1.0}
\begin{table*}[t!]
\centering
\caption{{\bf Overall Performance Comparison among TFLite, SNPE, and \projectname on Mobile DSP.} ``-'' means this model is not supported by the framework yet. OverT and OverS are the speedup of \projectname over TFLite, and SNPE, respectively.  \projectname's overall {\bf\em compilation time} for these models ranges from 5 minutes (WDSR-b) to 25 minutes (EfficientDet-d0).}
\label{tab:eva_performance_report}
\scriptsize
\begin{tabular}{lcc|ccc|c|c|c|cc}
     \toprule
     Model & Type & Task & \#MACS & \#Params & \#Operators & TFLite (ms) & SNPE (ms) & \projectname (ms) & OverT & OverS \\ 
     \hline
     MobileNet-V3    & 2D CNN      & Classification       & 0.22G & 5.5M  & 193 & 7.5   & 6.2  & {\bf 4.0} & {\bf 1.9} & {\bf 1.6} \\
     EfficientNet-b0 & 2D CNN      & Classification       & 0.40G & 4M    & 254 & 9.1   & 9.2  & {\bf 6.0} & {\bf 1.5} & {\bf 1.5} \\
     ResNet-50       & 2D CNN      & Classification       & 4.1G  & 25.5M & 140 & 13.9  & 11.6 & {\bf 7.1} & {\bf 2.0} & {\bf 1.6} \\
     FST             & 2D CNN      & Style transfer       & 161G  & 1.7M  & 64  & 935   & 870  & {\bf 211} & {\bf 4.4} & {\bf 4.1} \\
     CycleGAN        & GAN         & Image trans.    & 186G  & 11M   & 84  & 450   & 366  & {\bf 181} & {\bf 2.5} & {\bf 2.0} \\
     WDSR-b          & 2D CNN      & SR     & 11.5G & 22.2K & 32  & 400   & 137  & {\bf 66.7}& {\bf 6.0} & {\bf 2.1} \\
     EfficientDet-d0 & 2D CNN      & 2D obj. detect  & 2.6G  & 4.3M  & 822 & 62.8  & -    & {\bf 26}  & {\bf 2.4} & {\bf -}   \\
     PixOr           & 2D CNN      & 3D obj. detect  & 8.8G  & 2.1M  & 150 & 43    & 26.4 & {\bf 11.7}& {\bf 3.7} & {\bf 2.3} \\
     TinyBERT        & Transformer & NLP                  & 1.4G  & 4.7M  & 211 & -     & -    & {\bf 12.2}& {\bf -}   & {\bf -}   \\
     Conformer       & Transformer & Speech recog.   & 5.6G  & 1.2M  & 675 & -     & -    & {\bf 65}  & {\bf -}   & {\bf -}   \\
    \hline
     \multicolumn{9}{c}{Speedup (geometric mean)}                                                          & {\bf 2.8} & {\bf 2.1} \\
     \bottomrule
\end{tabular}
\end{table*}
\endgroup

Our \projectname is a general framework that applies not only to mobile CPU and GPU, but also applies to dedicated NPU/DSP devices and micro-controllers (MCUs) and achieves significant speedup. 

This part first evaluates the latency of \projectname by comparing it against two state-of-the-art frameworks, TFLite~\cite{tensorflowlite} and SNPE~\cite{snpe} on mobile DSP.
Table~\ref{tab:eva_performance_report} shows the comparison for 10 cutting-edge models on a Samsung Galaxy S20 (with Snapdragon 865 SoC~\cite{snapdragon865}) with Hexagon 698 DSP (with Vector eXtensions support). TFLite and SNPE do not support Transformer-based models. For the other 
8 models,  \projectname achieves 1.5$\times$ to 6.0$\times$, and 1.5$\times$ to 4.1$\times$ speedup over TFLite and SNPE, respectively. 
Table~\ref{tab:eva_performance_report} shows that
\projectname achieves the most speedup (6.0$\times$ over TFLite) on WDSR-b because the feature map shapes in WDSR vary significantly among different operators. This enables more benefits from our compiler optimizations (e.g., instruction selection and layout transformation optimizations).    
Particularly, \projectname  for the {\em first} time supports TinyBERT and Conformer executed on mobile DSP because it supports more operators than TFLite and SNPE, e.g., more variants of {\tt MatMul}, and {\tt Pow}. 
It also {\em the first time} achieves real-time execution on mobile DSP for EfficientDet-d0. 
We also compared several individual convolutional computation kernels with Halide~\cite{halide} and TVM~\cite{chen2018tvm}, as the end-to-end inference is not yet supported from these frameworks on the DSP chip. Specifically, first 8 unique {\tt Conv2D} operators in ResNet-50 are used. \projectname achieves $3-5\times$ speedup over Halide and TVM.

\begin{figure}[h]
\centering
\includegraphics[width=0.45\textwidth]{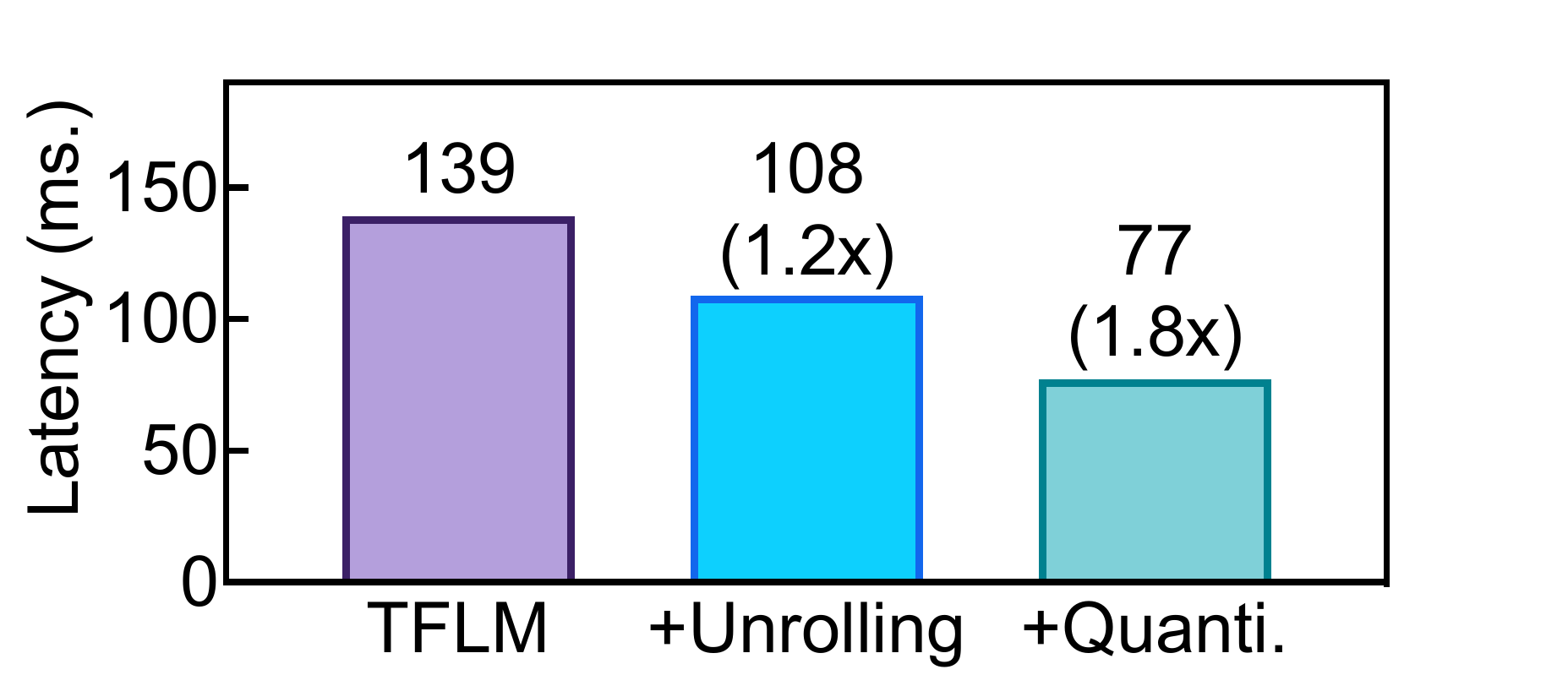}
\caption{Latency comparison between TFLM (TensorFlow Lite Micro) and XGen with optimizations (Unrolling and Optimized Quantization), respectively.}
\label{fig:latency}
\end{figure}

This part next compares \projectname with a state-of-the-art inference framework, TensorFlow Lite Micro (TFLM)~\cite{MLSYS2021_d2ddea18} on a popular Micro-controller Unit (MCU), STM32F469NI. 
TFLM leverages the high-performance ARM CMSIS-NN libraries~\cite{lai2018cmsis} for common DNN operations (e.g., convolution) to deliver optimized performance.
Figure~\ref{fig:latency} compares the inference latency of \projectname and TFLM on an optimized MobileNet-V2.
With compiler optimizations (e.g., loop unrolling that reduces the register spilling), \projectname can achieve $1.2\times$ speedup over TFLM.
With our further optimized quantization, \projectname can achieve $1.8\times$ speedup over TFLM.

\subsubsection{Benefits from AI-Aware Runtime} 

The benefits from the AI-aware runtime of CoCoPIE are demonstrated in the deployment of a complicated Level-4 autonomous driving (shown in Figure~\ref{fig:am}) on a low-end single-board device, NVIDIA Xavier Jetson. The current industry has regarded Level-4 autonomous driving possible only on high-end devices, such as NVIDIA Petegus (worth over \$10000). Jetson, worth \$700, has only a small fraction of the computing power of Petegus. Our experiments show that if we directly deploy Level-4 autonomous driving applications on Jetson, the application makes no progress at all, as Segment 1 in Table~\ref{tab:bigtable} shows. The reason is that the contentions for the limited computing resources by the many AI models in the application cause a deadlock. After we port the application to use the default Linux time-sharing scheduler, the deadlock is resolved but the 2D perception modules are about twice as slow as its required speed, as Segment 2 in Table~\ref{tab:bigtable} shows. The reason is that the many AI modules cause the 2D perception module difficult to get enough computing resource. Some computing units (DLA) are still underutilized but cannot be fully leveraged by the AI modules due to some mismatches between the model structures and the hardware constraints, as well as the limitations of the runtime scheduling. After equipping the device with CoCoPIE runtime, the application runs smoothly, meeting the real-time requirements completely, as Segment 5 in Table~\ref{tab:bigtable} shows. Segments 3 and 4 in Table~\ref{tab:bigtable} give the ablation studies on the benefits from each of the optimizations of the CoCoPIE runtime. 

\begin{table*}
\centering
\caption{Execution time (mean$\pm$ std)  of each module in the ADApp applications on Jetson AGX Xavier and the miss rates. {\small The $\infty$ represents timeout. The \texttt{miss rate} of a module is how often the module misses its expected latency (shown in the parentheses in the table header)---up to 10\% over is allowed to tolerate system noises. The column \texttt{Miss Rate} shows the miss rates of the most sluggish modules (whose times are prefixed with an $\ast$), that is, the modules with the largest miss rate in the application.}}\label{tab:bigtable}
\renewcommand{\arraystretch}{1.4}
\scriptsize
\begin{tabular}{|l||c|cc|c|c|c|c||c|}\hline
Application  & \multicolumn{7}{|c||}{Running Time of Each Module (ms) \textbf{[expected latency in brakets]}} & {\bf Miss Rate} \\ \cline{2-8}
& \emph{Sensing} & \emph{3D Percept} & \emph{2D Percept} & \emph{Localization} & \emph{Tracking} & \emph{Prediction} & \emph{Planning} &  \\ 
& [100ms] & [100ms] & [100ms] & [100ms] & [100ms] & [100ms] & [10ms] & \\ \hline
\multicolumn{9}{|l|}{1. Default ROSCH} \\ \hline
ADy288 & $8.8\pm1.0$ & * $\infty$ & * $\infty$ & * $\infty$ & * $\infty$ & * $\infty$ & $1.1\pm0.8$ & $100\%$\\
ADy416 & $8.5\pm0.7$ & * $\infty$ & * $\infty$ & * $\infty$ & * $\infty$ & * $\infty$ & $1.3\pm0.9$ & $100\%$ \\
ADy608 & $8.5\pm0.8$ & * $\infty$ & * $\infty$ & * $\infty$ & * $\infty$ & * $\infty$ & $1.1\pm0.7$ & $100\%$ \\
ADs288 & $9.0\pm0.8$ & * $\infty$ & * $\infty$ & * $\infty$ & * $\infty$ & * $\infty$ & $1.2\pm1.0$ & $100\%$ \\
ADs416 & $8.4\pm1.0$ & * $\infty$ & * $\infty$ & * $\infty$ & * $\infty$ & * $\infty$ & $1.3\pm0.6$ & $100\%$ \\
ADs608 & $8.5\pm0.9$ & * $\infty$ & * $\infty$ & * $\infty$ & * $\infty$ & * $\infty$ & $1.5\pm0.8$ & $100\%$ \\ \hline
\multicolumn{9}{|l|}{2. Default Linux Time Sharing} \\ \hline
ADy288& $14.3\pm5.2$ & $94.7\pm12.8$ & * $193.3\pm17.5$ & $89.5\pm30.5$ & $0.9\pm0.8$ & $0.4\pm1.0$ & $1.0\pm0.1$ & $100\%$ \\
ADy416 & $15.3\pm5.1$ & $90.2\pm12.0$ & * $167.6\pm12.7$ & $89.1\pm29.1$  & $0.9\pm0.7$  & $0.5\pm0.9$ & $1.1\pm0.2$  & $100\%$  \\
ADy08 & $14.8\pm4.8$ & $89.0\pm18.7$ & * $192.8\pm16.2$  & $91.5\pm31.2$ & $1.1\pm0.7$ & $0.4\pm0.9$ & $1.1\pm0.2$ & $100\%$ \\
ADs288 & $14.3\pm5.0$ & $95.6\pm13.7$ & * $195.2\pm18.2$  & $88.7\pm28.9$ & $1.0\pm0.8$ & $0.4\pm1.0$ & $1.1\pm0.1$ & $100\%$ \\
ADs416 & $14.8\pm4.8$ & $91.3\pm13.4$ & * $168.8\pm13.3$ & $90.1\pm30.2$ & $1.1\pm0.9$ & $0.5\pm1.0$ & $1.1\pm0.0$ & $100\%$ \\
ADs608 & $14.7\pm4.9$ & $90.6\pm19.2$ & * $194.2\pm17.7$ & $91.2\pm29.3$ & $0.9\pm0.6$ & $0.4\pm1.1$ & $1.1\pm0.1$ & $100\%$ \\ \hline
\multicolumn{9}{|l|}{3. Just-In-Time (JIT) Priority Adjustment} \\ \hline
ADy288 & $8.5\pm0.9$ & $94.6\pm13.4$ & * $194.7\pm16.3$ & $43.5\pm10.2$ & $1.0\pm0.7$ & $0.6\pm1.1$ & $1.2\pm0.4$ & $100\%$ \\
ADy416 & $8.4\pm1.0$ & $91.7\pm11.2$ & * $166.8\pm11.4$ & $45.3\pm11.3$ & $0.8\pm1.0$ & $0.6\pm1.1$ & $1.0\pm0.4$ & $100\%$ \\
ADy608 & $8.7\pm0.7$ & $88.9\pm17.6$ & * $190.9\pm17.9$ & $47.2\pm9.9$ & $1.1\pm0.6$ & $0.5\pm0.9$ & $1.0\pm0.5$ & $100\%$ \\
ADs288 & $8.9\pm0.9$ & $96.1\pm12.7$ & * $195.9\pm17.3$ & $46.9\pm12.1$ & $0.9\pm0.9$ & $0.6\pm1.0$ & $1.0\pm0.4$ & $100\%$ \\
ADs416 & $9.0\pm0.7$ & $92.8\pm11.4$ & * $169.7\pm13.3$ & $48.1\pm11.9$ & $1.0\pm0.5$ & $0.5\pm1.1$ & $1.3\pm0.4$ & $100\%$ \\
ADs608 & $8.7\pm0.9$ & $91.2\pm20.3$ & * $194.8\pm16.9$ & $44.7\pm10.9$ & $0.8\pm1.0$ & $0.4\pm1.2$ & $1.2\pm0.2$ & $100\%$ \\ \hline
\multicolumn{9}{|l|}{4. JIT Adjustment + Migration to Accelerators} \\ \hline
ADy288 & $8.7\pm0.8$ & $123.8\pm18.5$ & * $225.6\pm5.0$ & $43.0\pm9.9$ & $0.9\pm1.0$ & $0.5\pm0.9$ & $1.0\pm0.6$ & $100\%$ \\
ADy416 & $8.8\pm1.1$ & $128.7\pm12.1$ & * $177.6\pm3.3$ & $46.7\pm10.5$ & $1.0\pm0.8$ & $0.5\pm1.0$ & $1.2\pm0.3$  & $100\%$ \\
ADy608 & $9.1\pm0.9$ & $144.3\pm8.1$ & * $171.8\pm3.0$ & $48.5\pm9.8$ & $1.0\pm0.8$ & $0.6\pm1.1$ & $1.2\pm0.3$ & $100\%$ \\
ADs288 & $9.0\pm0.8$ & $125.6\pm17.1$ & * $225.6\pm6.3$ & $47.3\pm11.3$ & $1.1\pm0.7$ & $0.4\pm1.1$ & $1.3\pm0.2$ & $100\%$ \\
ADs416 & $8.8\pm1.1$ & $130.5\pm13.2$ & * $180.1\pm4.7$ & $47.6\pm9.8$ & $0.9\pm0.9$ & $0.7\pm1.2$ & $1.5\pm0.2$ & $100\%$ \\
ADs608 & $8.6\pm0.9$ & $147.2\pm9.2$ & * $174.3\pm4.5$ & $47.6\pm10.4$ & $0.9\pm0.9$ & $0.6\pm1.2$ & $1.2\pm0.5$ & $100\%$ \\ \hline
\multicolumn{9}{|l|}{5. JIT Adjustment + Migration to Accelerators + Model-Schedule Co-optimization} \\ \hline
ADy288 & $8.4\pm1.2$ & * $89.0\pm15.3$ & $95.6\pm5.1$ & $46.3\pm9.8$ & $0.9\pm0.9$ & $0.7\pm0.9$ & $1.0\pm0.4$ & $0\%$ \\
ADy416 & $9.0\pm0.9$ & $72.0\pm9.0$ & $88.1\pm4.3$ & $44.9\pm10.7$ & $1.0\pm0.8$ & $0.6\pm0.9$ & $1.3\pm0.2$  & $0\%$ \\
ADy608 & $8.8\pm1.2$ & $80.8\pm10.6$ & * $98.1\pm5.0$ & $46.4\pm11.0$ & $1.0\pm0.7$ & $0.4\pm1.1$ & $1.1\pm0.3$ & $0\%$ \\
ADs288 & $9.0\pm1.1$ & * $92.0\pm14.3$ & $96.4\pm5.4$ & $45.8\pm10.1$ & $1.1\pm0.7$ & $0.4\pm1.1$ & $1.2\pm0.6$ & $0\%$ \\
ADs416 & $8.9\pm0.8$ & $74.2\pm9.3$ & $90.0\pm4.2$ & $47.2\pm10.0$ & $1.0\pm0.8$ & $0.5\pm0.9$ & $1.2\pm0.2$ & $0\%$ \\
ADs608 & $8.8\pm1.0$ & $83.7\pm9.7$ & * $100.1\pm4.4$ & $46.8\pm9.9$ & $0.9\pm0.7$ & $0.7\pm0.8$ & $1.3\pm0.8$ & $0\%$ \\ \hline
\end{tabular}
\end{table*}

\section{XGen as a Product}

\begin{figure}
    \centering
    \includegraphics[width=.99\textwidth]{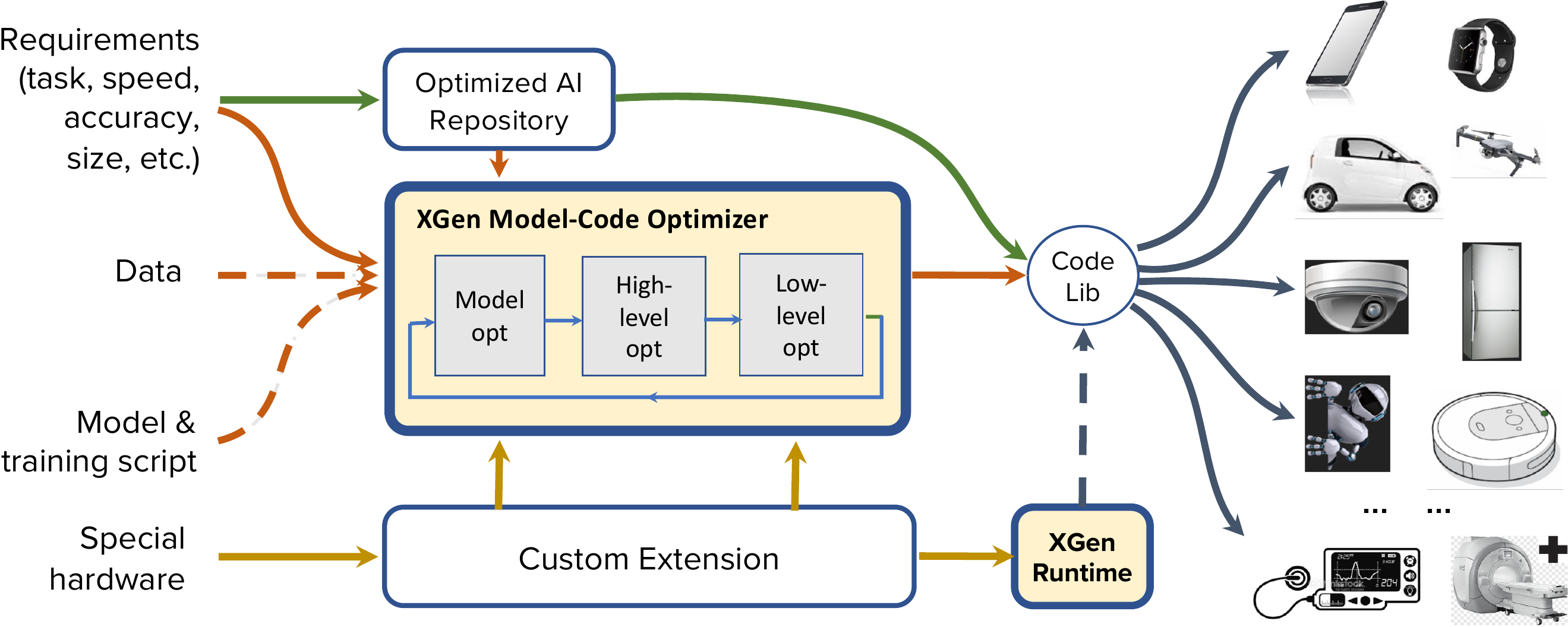}
    \caption{The service modes and paths of XGen. The paths of three colors (green, red, yellow) indicate the three service paths offered by XGen. Broken lines indicate optional components.}
    \label{fig:paths}
\end{figure}

This section describes the forms of XGen as a product and its usage. In our design, XGen has two forms. One is the form of Software-As-A-Service (SAAS) on the cloud, the other is a standalone software package installable on a single machine or a cluster. The first form makes it easy to use by customers, who can start using XGen immediately without installation while paying only for use. It also makes it possible for XGen to serve many customers at the same time in a scalable manner. The second form helps meet the needs of customers who have difficulties or concerns in uploading their data or DNN models to a third-party cloud. By installing XGen in their own cluster or datacenter, they can use it inside their organization. 

In either form, XGen can be used in several ways to meet the needs in various scenarios, as shown in Figure~\ref{fig:paths}. 
\begin{itemize}
    \item {\em Scenario I:} A customer needs a common AI capability with a certain requirement (speed on some common devices, accuracy, size, etc.) that is met by some of the models XGen has produced before. The customer has no particular requirements on what DNN models or datasets to use.\\
    {\em Usage I:} The customer provides her requirements through the XGen interface, and XGen immediately returns the AI capability it has already stored in its repository that meets the requirements. The top green path in Figure~\ref{fig:paths} shows this service path.
    \item {\em Scenario II:} Similar to Scenario I except that the requirement (speed, accuracy, size, etc.) from the customer is not met by any of the models XGen has produced before. \\
    {\em Usage II:} The customer provides her requirements through the XGen interface, and XGen identifies promising base DNN models and then conducts its optimizations to generate efficient code that meets the user's requirements. The solid red path in Figure~\ref{fig:paths} shows this service path.
    \item {\em Scenario III:} Similar to Scenario II except that the customer has her own dataset and/or DNN model. This scenario arises typically when the customer needs a special AI capability that requires custom model training and optimization.\\
    {\em Usage III:} The customer provides her requirements as well as her dataset and/or DNN model and the model's training script that have been prepared based on XGen guidelines. XGen can seamlessly integrate the customer's training script into its workflow, optimizes the DNN model in a way similar to Usage II, and produces the code that meets the customer's required accuracy on her dataset and her other specified requirements. The solid plus broken red lines in Figure~\ref{fig:paths} show this service path.
    \item {\em Scenario IV:} The customer wants to use XGen to optimize DNNs for a special hardware (e.g., a new AI accelerator). \\
    {\em Usage IV:} XGen is designed to fit the features of common processors (e.g., GPU, CPU, DSP). It is at the same time extensible. Custom extensions can be easily built into XGen to support other special processors. XGen guarantees this extensibility by designing a general Intermediate Representation (IR) that captures the key DNN operator features and modern AI accelerator design trends, such as increasingly powerful SIMD and the integration of SIMD and VLIW features. This IR enables fast/low-cost new back-end ISA support.
\end{itemize}

In any of the usage scenario, the execution of the optimized DNN can benefit from XGen runtime if it is installed on the target device. The use of XGen runtime is optional. Without it, the optimized DNN can already run efficiently; with it, the efficiency can be better assured in a multi-tenant resource constrained environment. 


\section{Example Use Cases}

\begin{figure}
    \centering
    \fbox{\includegraphics[width=.6\textwidth]{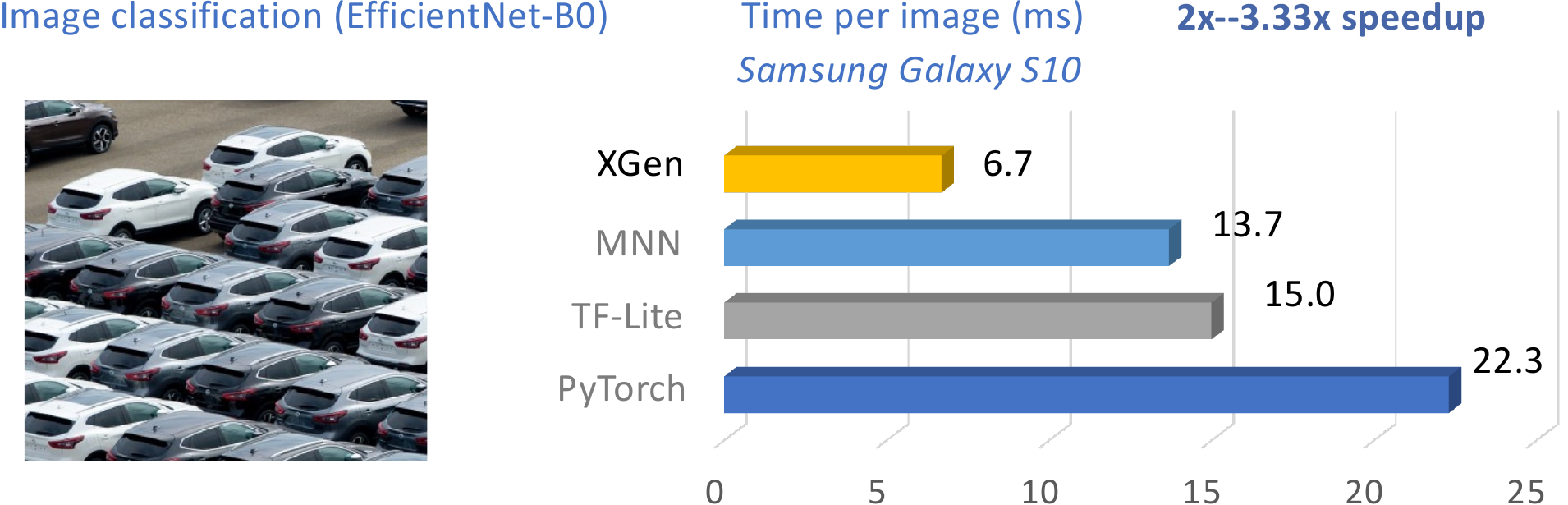}}\\[4mm]
    \fbox{\includegraphics[width=.6\textwidth]{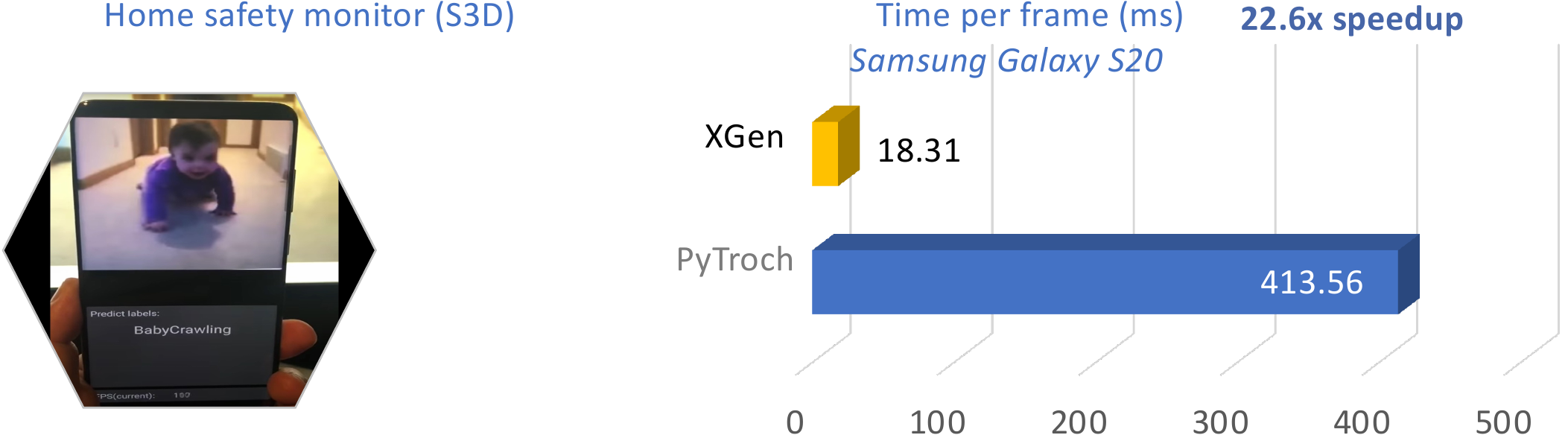}}\\[4mm]
    \fbox{\includegraphics[width=.6\textwidth]{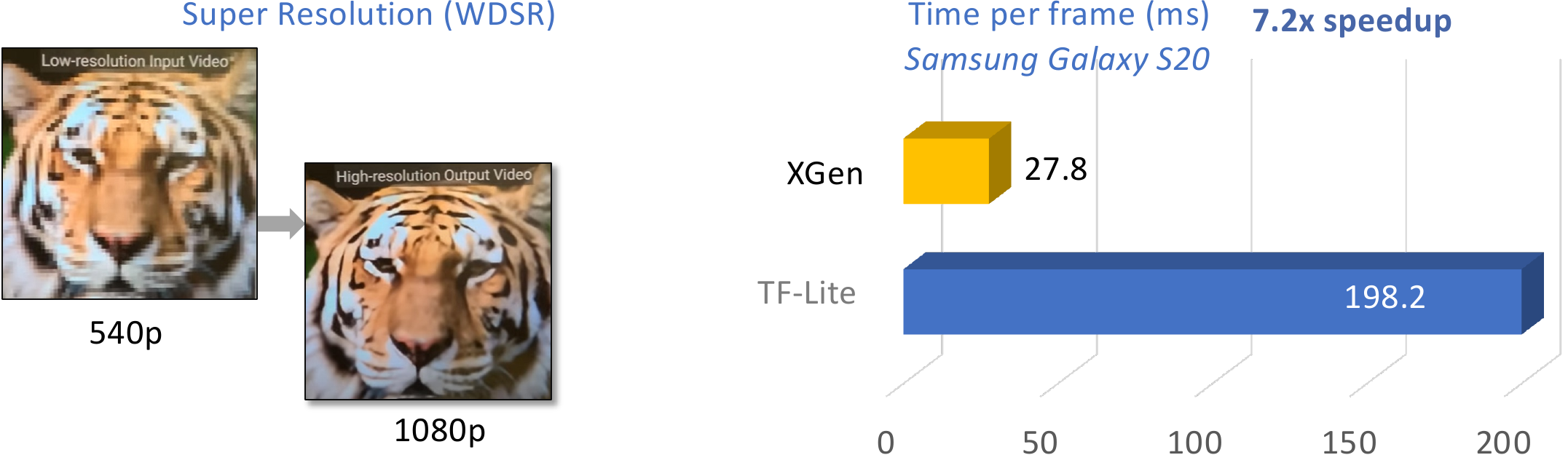}}
    \caption{Three use cases and the comparisons with other DNN frameworks.}
    \label{fig:useCases}
\end{figure}

This section draws on three example use cases to explain the practical impact of XGen. The three use cases are car classification, home safety monitor, and super resolution. Figure~\ref{fig:useCases} shows each of them and the performance measurements on Samsugn Galaxy S10 and S20 cellphones. We next provide a brief explanation of each of them.

\begin{itemize}
    \item {\em Use case I: Car classification.} Company A needs to develop a smartphone app for its customers with which the customers can recognize the models of vehicles in real-time. It is a typical image classification problem, one of the most common uses of DNN. For its popularity, many efforts have been spent by the industry to optimize the speed. But even with that, XGen still brings 2$\times$-3.33$\times$ speedups over the results from the mainstream DNN frameworks, PyTorch, TF-Lite, MNN, while keeping the classification accuracy unchanged. The speedups come from the 
    significant reduction of computations by the model pruning and the enhanced parallelism on GPU brought by the pattern-aware data layout and other optimizations.   
    
    \item {\em Use case II: Home safety monitor.} Company B wants to develop a smart app which can recognize the activity of babies or elders at real-time and send out alarms when there is safety risks. The core of the app is a DNN for activity recognition. Compared to the DNN models used in image classification, this task uses a more complex DNN model, S3D, which considers both spatial and temporal dimensions of the inputs. In our experimented existing DNN frameworks, only PyTorch was able to produce code that successfully ran on the device. XGen shows a 22.6$\times$ speedup over PyTorch, while giving the same accuracy. The larger size and complexity of the DNN model provide even more room for the full-stack cooperative optimizations of XGen to take effect. 
    The dramatic acceleration by XGen is a game changer: With a speed of 18.31ms per frame, it for the fist time enables {\em real-time} activity recognition on a smartphone, which makes the real-time safety monitoring possible.
    
    \item {\em Use case III: Super resolution.} Company C is a video content provider. It would like to develop a video player that can automatically upscale an image or a video to a higher resolution in real time. One of the purposes is to improve their user experience when users watch a streaming video in a unstable network condition. The other purpose is to save the network bandwidth consumption and hence cost: If the upscaling can take place on the user's device, a lower resolution stream can be transferred to the user without compromising her experience. The used DNN model is WDSR. TF-Lite is the only existing DNN framework working on this task in our experiments. Without model pruning, {\em XGen} can already outperform TF-Lite by 1.9$\times$, thanks to its advanced optimizations (e.g., operator replacement, SIMD optimizations, etc.). When combined with pattern-based pruning, XGen produces $3.7\times$ extra speedups, yielding overall 7.2$\times$ speedups over TF-Lite. The numbers of frames per second increase from 5 to 36, for the first time making real-time super resolution on mobile devices a reality. 
\end{itemize}

 Please see video demos of more use cases at the CoCoPIE YouTube channel\footnote{\url{ http://www.youtube.com/channel/UCCKVDtg2eheRTEuqIJ5cD8A/}} and Bilibili channel\footnote{\url{https://space.bilibili.com/573588276?from=search&seid=11881710196887435131}}.

\newpage
\bibliographystyle{unsrt}
\bibliography{all,pldi}






\end{document}